\documentclass[sigconf]{acmart}

\AtBeginDocument{%
  }

\usepackage{times}
\usepackage{balance}
\usepackage{latexsym}
\usepackage{booktabs}
\usepackage{colortbl}
\usepackage{xcolor}
\usepackage{multirow}
\usepackage{geometry}
\usepackage{amsmath}
\usepackage{bbm}
\usepackage{bm}
\usepackage{color}
\usepackage{multirow}
\usepackage{array}
\usepackage{booktabs}
\usepackage{diagbox}
\usepackage{bbding}
\usepackage{colortbl}
\usepackage{tabularx}
\usepackage{makecell}
\usepackage{hyperref}
\usepackage{subfigure}

\usepackage{amsmath, amssymb, overpic, textpos}
\usepackage{algorithmicx}
\usepackage{algpseudocode}
\usepackage{listings}
\usepackage{multicol}
\usepackage{xspace}
\usepackage{wrapfig}
\usepackage{graphicx}
\usepackage{graphbox}
\usepackage{arydshln}
\usepackage{algorithm}
\usepackage{adjustbox}
\usepackage{kotex}
\usepackage{caption}
\usepackage{setspace} 
\usepackage{epigraph}
\usepackage{tcolorbox}
\acmISBN{978-1-4503-XXXX-X/2018/06}
\usepackage{tcolorbox}
\newcommand{\ds}{{ADS-Edit}\xspace}
\definecolor{mygray}{gray}{.8}
\newcommand{\GG}{\cellcolor{mygray}}

\copyrightyear{2025}
\acmYear{2025}
\setcopyright{acmlicensed}\acmConference[MM '25]{Proceedings of the 33rd ACM International Conference on Multimedia}{October 27--31, 2025}{Dublin, Ireland}
\acmBooktitle{Proceedings of the 33rd ACM International Conference on Multimedia (MM '25), October 27--31, 2025, Dublin, Ireland}
\acmDOI{10.1145/3746027.3758237}
\acmISBN{979-8-4007-2035-2/2025/10}

\begin{document}

\title{ADS-Edit: A Multimodal Knowledge Editing Dataset for Autonomous Driving Systems}

\author{Chenxi Wang}
\authornote{Both authors contributed equally to this research.}
\email{sunnywcx@zju.edu.cn}
\author{Jizhan Fang}
\authornotemark[1]
\email{fangjizhan@zju.edu.cn}
\affiliation{%
  \institution{Zhejiang University,\\Zhejiang University - Ant \\Group Joint Laboratory of Knowledge Graph}
  \city{Hangzhou}
  \country{China}}

\author{Xiang Chen}
\authornotemark[1]
\affiliation{%
  \institution{Zhejiang University,\\Zhejiang University - Ant \\Group Joint Laboratory of Knowledge Graph}
  \city{Hangzhou}
  \country{China}}
\email{xiang_chen@zju.edu.cn}

\author{Bozhong Tian}
\email{tbozhong@zju.edu.cn}
\author{Ziwen Xu}
\email{ziwen.xu@zju.edu.cn}
\affiliation{%
  \institution{Zhejiang University,\\Zhejiang University - Ant \\Group Joint Laboratory of Knowledge Graph}
  \city{Hangzhou}
  \country{China}}

\author{Huajun Chen}
\affiliation{%
  \institution{Zhejiang University,\\Zhejiang University - Ant \\Group Joint Laboratory of Knowledge Graph}
  \city{Hangzhou}
  \country{China}}
\email{huajunsir@zju.edu.cn}

\author{Ningyu Zhang}
\authornote{Corresponding author.}
\affiliation{%
  \institution{Zhejiang University,\\Zhejiang University - Ant \\Group Joint Laboratory of Knowledge Graph}
  \city{Hangzhou}
  \country{China}}
\email{zhangningyu@zju.edu.cn}

\begin{abstract}
Recent advancements in Large Multimodal Models (LMMs) have shown promise in Autonomous Driving Systems (ADS). However, their direct application to ADS is hindered by challenges such as misunderstanding of traffic knowledge, complex road conditions, and diverse states of vehicle. To address these challenges, we propose the use of Knowledge Editing, which enables targeted modifications to a model's behavior without the need for full retraining. Meanwhile, we introduce \ds, a multimodal knowledge editing dataset specifically designed for ADS, which includes various real-world scenarios, multiple data types, and comprehensive evaluation metrics. We conduct comprehensive experiments and derive several interesting conclusions. We hope that our work will contribute to the further advancement of knowledge editing applications in the field of autonomous driving\footnote{Code and data are available in \url{https://github.com/zjunlp/EasyEdit/blob/main/examples/ADSEdit.md}.}.
\end{abstract}





\maketitle

\section{Introduction}
% intro 提纲
% 段落1：多模态大模型发展迅猛，在自动驾驶领域也有应用。,deepseekvl2,nvila,qvq,
The recent Large Multimodal Models (LMMs) \citep{qwen2vl,llavaonevision,qwen2.5-VL} have significantly enhanced capabilities in video understanding and multimodal reasoning. 
As a key foundation, LMMs have also found initial applications in Autonomous Driving Systems (ADS) \citep{openemma,drivelm,drivemm,calmm}. 
However, the direct application of LMMs in ADS yields suboptimal results.

\begin{figure}[t]
    \centering
    \includegraphics[width=1\linewidth]{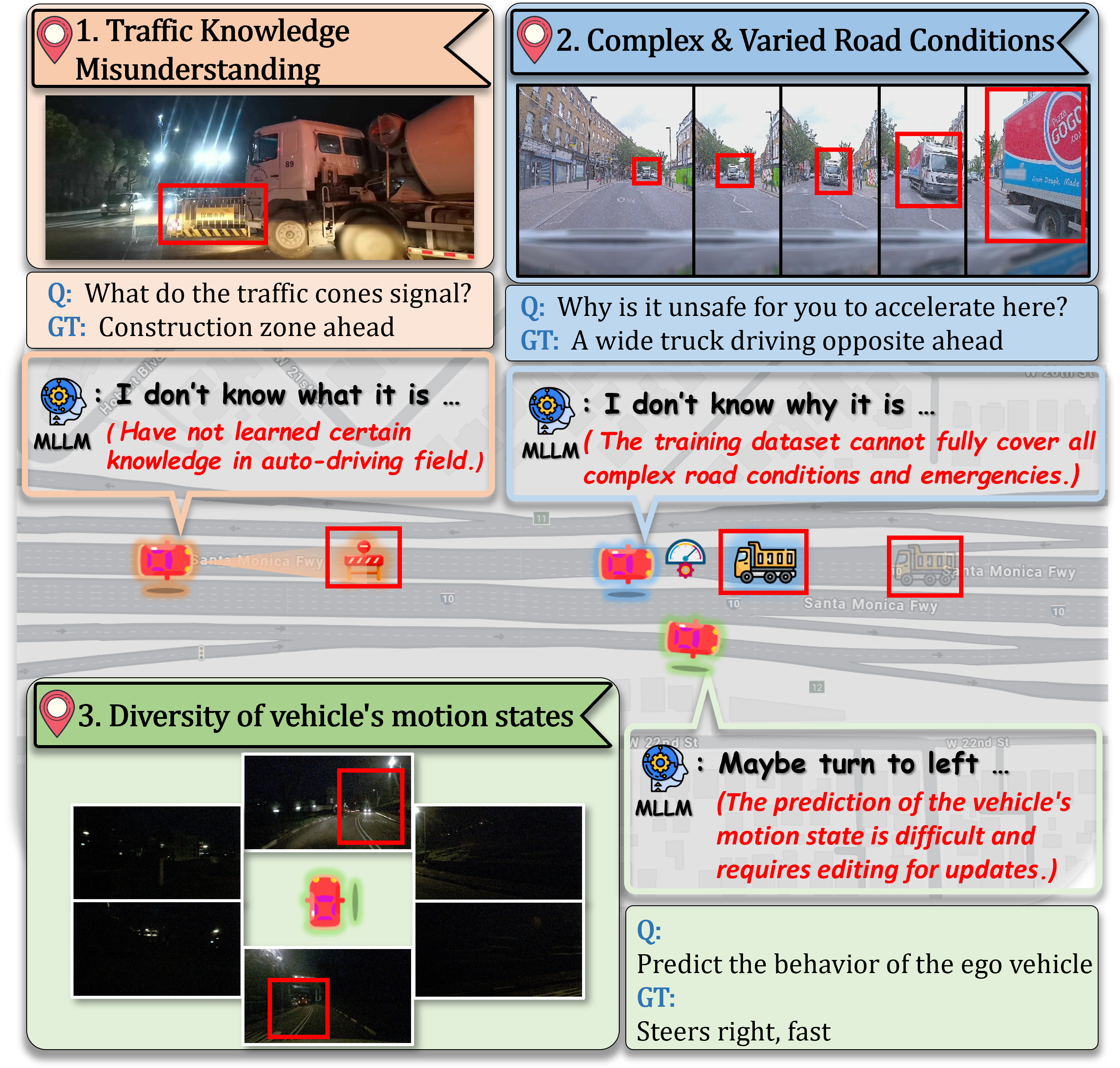}
    \caption{Direct application of LMMs in Autonomous Driving Systems faces several challenges, including the misunderstanding of traffic knowledge, the complex and varied road conditions, and the diversity of vehicle's motion states. Knowledge Editing that enables efficient, continuous, and precise updates to knowledge can effectively address these challenges.
    }
    \label{fig:motivation}
\end{figure}

% 段落2：现有的多模态大模型在3个场景下的应用存在局限行，需要编辑技术来快速更新。
% However, while general-purpose LMMs excel in common scenarios, their direct application to ADS reveals several challenges, as illustrated in Figure~\ref{fig:motivation}:
As shown in Figure~\ref{fig:motivation}, the reasons for failure can be attributed to the following factors:
\textbf{1) Traffic knowledge misunderstanding:} General models misunderstand traffic knowledge leading to suboptimal performance in tasks.
\textbf{(2) Complex and varied road condition:} Real-world driving scenarios are highly variable, with training datasets often failing to cover edge cases. 
\textbf{(3) Diversity of vehicle motion states:} Current LMMs struggle to predict unknown and highly dynamic vehicle motion states.
These challenges require the model to have the capability of updating knowledge in real-time and continuously.

% \textbf{(1) Lack Domain Knowledg:} General models lack embedded domain knowledge 
% leading to suboptimal performance in tasks. \textbf{(2) Complex and Varied Road Condition:} Real-world driving scenarios are highly variable, with training datasets often failing to cover edge cases. \textbf{(3) Diversity of vehicle motion states:} Current LMMs struggle to predict unknown and highly dynamic vehicle motion states.

To address these challenges, we propose the use of Knowledge Editing, which enables targeted modifications to the model’s behavior, eschewing the substantial computational demands associated with full retraining.
% without the computational burden of full retraining. 
Unlike traditional fine-tuning, which risks catastrophic forgetting and demands extensive resources, \textbf{Knowledge Editing facilitate rapid knowledge updates by selectively altering parameters associated with specific factual or contextual knowledge} \citep{yao2023editing,gupta2024model,comprehensive,DBLP:journals/corr/abs-2410-02355,youssef2025position,jiang2025anyedit,liu2024codeupdatearena,yao2025cake}.
While preliminary explorations of Knowledge Editing in the area of LMMs have predominantly concentrated on the manipulation of multimodal common knowledge \citep{cheng2023mmedit, comprehendedit} and multimodal factual knowledge \citep{mcmke, vlkeb}, we pioneer the editing of multimodal domain-specific knowledge in the field of autonomous driving.
% Although Knowledge Editing has been preliminarily explored in the area of LMMs, existing efforts have primarily focused on editing multimodal common knowledge \citep{cheng2023mmedit,comprehendedit} and multimodal factual knowledge \citep{mcmke,vlkeb}. To address this limitation, we pioneer the editing of multimodal domain-specific knowledge in the field of autonomous driving.

Following the principle of leveraging knowledge editing to address the challenges of autonomous driving system, we propose the \textbf{\ds} benchmark. 
This benchmark encompasses three real-world scenarios: perception, understanding, and decision making.
Additionally, it incorporates three types of data: video, multi-view images, and single image.
Furthermore, we establish comprehensive evaluation metrics for knowledge editing in autonomous driving scenarios.

We evaluate four commonly used Knowledge Editing baselines under both single editing and lifelong editing scenarios, such as Prompt, AdaLora \citep{adalora}, GRACE \citep{grace}, and WISE \citep{wise}.
Through our analysis, we obtained a series of interesting findings, including the universality of knowledge editing methods in updating knowledge across various scenarios, their ability to balance editing effectiveness and processing speed, particularly in the context of video data, and the remaining limitations in locality.  
Moreover, in real-world evaluation scenarios, it is evident that current editing techniques require further refinement and enhancement to meet the demands of autonomous driving systems.
% Thirdly, through our analysis, we obtained a series of interesting findings, including the universality of knowledge editing methods in updating knowledge across various scenarios, their ability to balance editing effectiveness and processing speed, particularly in the context of video data, and the remaining limitations in locality. 
% Furthermore, in real-world evaluation scenarios, we observe that current editing methods still have substantial room for improvement when applied to autonomous driving systems.

In general, we conclude our contributions as:
\begin{itemize}
    \item 
We are the first to attempt Knowledge Editing on multimodal domain knowledge data, specifically the ADS and effectively addresses the current challenges faced by LMMs when directly applied to ADS.

   \item 
We unveil \textbf{\ds}, a Knowledge Editing dataset specifically designed for the ADS. The dataset includes three common types of visual data and encompasses data that evaluates various model capabilities.
 
    \item 
We test four Knowledge Editing baselines under both single editing and lifelong editing settings and analyze several interesting results.
\end{itemize}

\section{Background}

LMMs have been applied in autonomous driving, typically after being trained on carefully curated driving datasets.
However, when faced with unfamiliar driving scenarios, such as new traffic regulations or predicting driver behavior during traffic congestion, the reliability of LMMs' decision making can significantly degrade.
Furthermore, LMMs struggle to maintain consistent performance when encountering sudden changes in road conditions, such as shifts in weather or unexpected traffic accidents. 
Finally, for continuously collected driving data, the challenge of updating knowledge in a timely and effective manner remains unresolved for LMMs. 
Overall, there is \textbf{a critical need for a framework that enables LMMs to rapidly and sequentially update knowledges when applied to autonomous driving.}

% We propose leveraging knowledge editing techniques to address challenges in autonomous driving scenarios, defined as follows: Let $f_\theta$ be a LMM and $f_{\theta e}$ be an edited model. Given the user's inputs $x_e$ (Includes text $t_e$ and autonomous driving's multimodal inputs $m_e$, such as image or videos) and the editing target $y_e$, the edited model is expected to modify LMM's outputs within the editing scope to match the editing target, while preserving the original model's output for the inputs outside the editing scope. The specific definition is as follows:
\begin{figure}[ht!]
    \centering
    \includegraphics[width=0.7\linewidth]{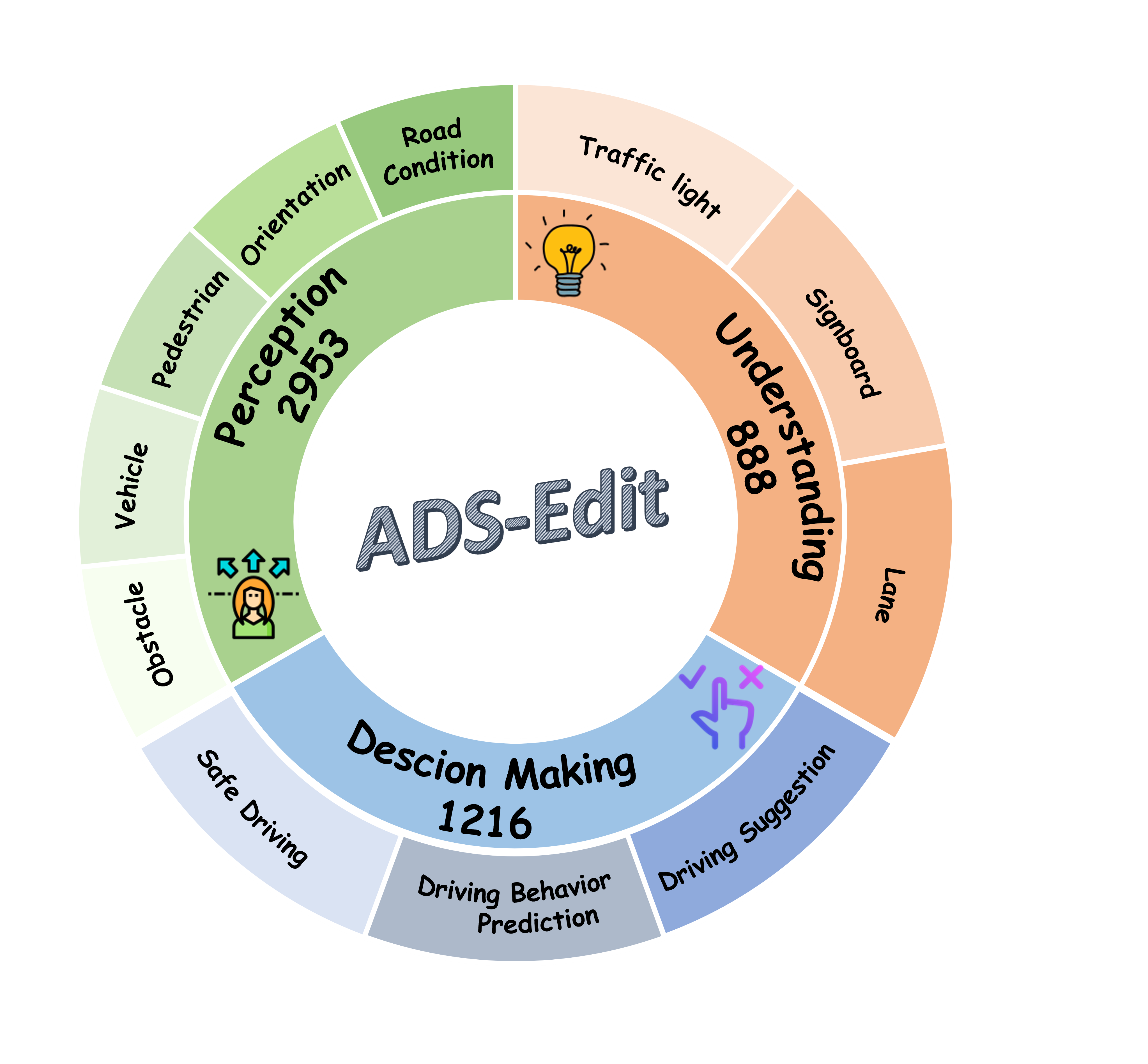}
    \caption{The statistics of scenario types for \ds.
    }
    \label{fig:scenario}
\end{figure}

We propose leveraging knowledge editing techniques to address the challenges encountered in autonomous driving scenarios. Formally, let $f_\theta$ denote a multimodal large model (LMM), and let $f_{\theta_e}$ represent the edited model. Given a user input $x_e$, consisting of textual input $t_e$ and multimodal signals $m_e$ from autonomous driving (e.g., images or videos), along with a desired editing target $y_e$, the goal is to update the model such that $f_{\theta_e}(x_e) \approx y_e$.
The edited model should modify the output of the original LMM within the intended editing scope while preserving its original behavior on inputs outside that scope. The formal definition is as follows:

\begin{equation}
f_{\theta e}(x) = 
\begin{cases} 
y_e & \text{if } x \in I(x_e, y_e) \\
f_{\theta}(x) & \text{if } x \notin I(x_e, y_e)
\end{cases}
\label{eq:def}
\end{equation}
where $x=(t,m)$ and $I(\cdot)$ means in-scope of editing inputs.

% Knowledge editing techniques provide robust frameworks to address these challenges, enabling efficiently updates to models without high-cost retraining. 
% GRACE \cite{grace} employs a discrete key-value codebook to cache edits as latent space mappings, preserving original model weights while enabling scalable sequential editing via a retrieval-based mechanism. 
% % This approach ensures strong locality and minimal interference with unrelated inputs.
% To bridge the gap between parametric and non-parametric memories, WISE \cite{wise} designs a dual-memory architecture. By maintaining a main memory for pretrained knowledge and a side memory for edited updates, WISE leverages a router to dynamically select between memories during inference. 
% % This approach preserves generalization by retaining parametric knowledge while enabling reliable, localized edits through side memory sharding and merging techniques.
% These innovations are particularly suited to autonomous driving, where real-time adaptation to road conditions (e.g., sudden weather changes) and motion prediction refinement demand both specificity and scalability. 
% By integrating Knowledge Editing’s efficient and flexibility, future systems could enable LMMs to dynamically assimilate domain-specific knowledge while preserving core driving competencies, ultimately advancing safer and more adaptive autonomous systems.

\begin{figure*}[ht]
    \centering
    \includegraphics[width=1\textwidth]{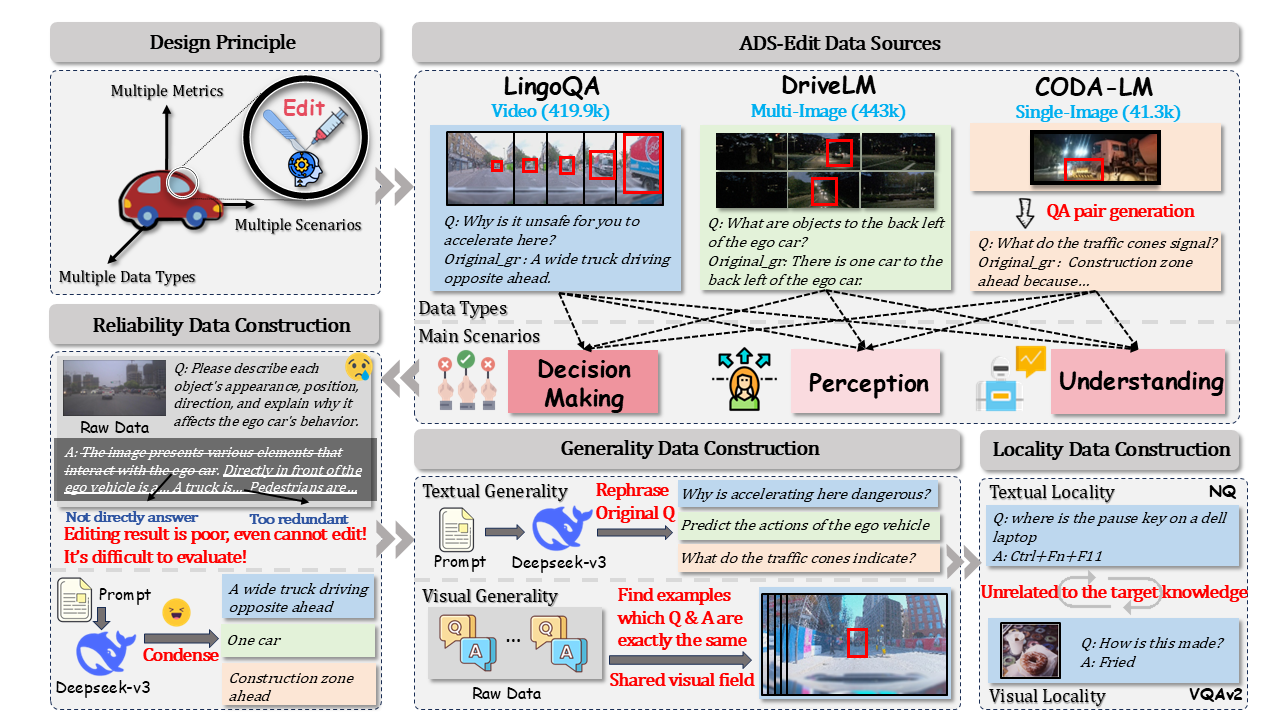}
    \caption{The overview of \textbf{\ds} construction pipeline.}
    \label{fig:overview}
\end{figure*}

\section{Bechmark Construction}

% 参考openreview上的建议改 尤其是 那几个ai人工审查+zny的建议 按场景来描述 别按老套的风格描述

\subsection{Desgin Principle}

% 重新整理设计思路：统一名称data types, task types

% 1 按实际需求，场景切分  2 数据都是可知识编辑的（以前很多不太小） 3 各个维度都测 全面

%1 基线

% 可能存在的问题：如单图不可能具有时序预测的数据。

% 任务类型介绍一段话 每一种类型是为了干什么的
% 多模态输入数据类型一段话 每种类型来源于哪

% To thoroughly evaluate the model's capabilities in autonomous driving scenario, we categorize the data requirements for multiple task types, including perception, understanding and decision making.
% 这段不是很地道
% To comprehensively evaluate the performance of knowledge editing methods in autonomous driving scenarios, we categorize data requirements into multiple task types, including perception, understanding, and decision making.
% Additionally, based on the type of visual input, the data are categorized into \textbf{Video}, \textbf{Multi-view images}, and \textbf{Single image}.
% To systematically evaluate the efficacy of knowledge editing methods for LMMs in autonomous driving applications, we propose a tri-axis design principle.
To construct a comprehensive benchmark, we propose a tri-axis design principle.
This principle organizes evaluation requirements into scenario types: perception, understanding, and decision making, which progressively assess LMMs’ capabilities from basic visual recognition to complex behavioral reasoning.
Concurrently, we categorize input modalities into data types: video, multi-view images, and single image.
Multiple metrics are designed to evaluate knowledge editing methods, such as reliability, generality and locality.

\paragraph{‌Scenario‌ Type.} \textit{Perception} scenario evaluates LMMs' basic visual perception capabilities, such as obstacle detection and vehicle recognition.
\textit{Understanding} scenario requires the model to comprehend domain-specific knowledge of autonomous driving, such as traffic rules, beyond basic perception capabilities.
\textit{Decision Making} scenario presents a greater challenge, as it requires the model to integrate perception and understanding capabilities to make informed decisions about future driving behaviors.
The statistics of scenario types is shown in Figure~\ref{fig:scenario}.

\begin{table}[h]
\centering
\small
\resizebox{0.8\columnwidth}{!}{
\resizebox{0.5\textwidth}{!}{% <------ Don't forget this %
\begin{tabular}{ccccc}
\toprule
& \bf Video & \bf Multi-view & \bf Single & \bf All  \\ 
\midrule
% \rule{0pt}{15pt}
Train & 1,926 & 960 & 1,093 & 3,979 \\
Test & 481 & 239 & 358 &  1,078 \\
% \rule{0pt}{15pt}
\bottomrule
\end{tabular}
}%
}
\caption{
Statistical information of \ds data types and dataset splits for training and testing.
}
\label{tab:static}
\end{table}

% Additionally, visual data types are classified into video, multiple images, and single image formats.
\paragraph{Data Type.} \textit{Video} requires the model to possess the ability to evaluate temporal changes in images. 
Specifically, the model must be capable of understanding the three dimensions of image width, height, and time.
\textit{Multi-view images} are designed to assess the model's capability when provided with sensor images from multiple viewpoints on the vehicle.
\textit{Single image} primarily tests the model's fundamental perception abilities, such as object recognition and spatial relationship understanding. We will organize these diverse data types to construct a comprehensive benchmark.
The statistics of different types of data is shown in Table~\ref{tab:static}.

% 创新写法 写多模态一点+可以写持续编辑以及单次编辑的公式
\paragraph{Metrics.}
\textit{Reliability} is to evaluate the success of behavioral modification in the target driving scenarios.
\textit{Generality} evaluates the generalization scores when facing similar autonomous driving scenarios.
\textit{Locality} measures whether methods alter unrelated knowledge after updating with domain-specific knowledge of autonomous driving.
% As for the comprehensive evaluation of Autonomous Driving System knowledge editing, inspired by previous studies \citep{yao2023editing,cheng2023mmedit}, we propose the use of metrics including \textbf{Reliability}, \textbf{Generality}, and \textbf{Locality}.

\subsection{Data Collection}
Inspired by knowledge editing applications in unimodal settings, the constructed \textbf{\ds} dataset is composed of visual question answering data.
The construction process of the ADS dataset is illustrated in Figure~\ref{fig:overview}.

\subsubsection{Data Preprocess.}
We select three well-known datasets of the autonomous driving system: LingoQA \citep{lingoqa}, DriveLM dataset \citep{drivelm}, and CODA-LM dataset \citep{codalm} as raw data. 
These datasets encompass a variety of autonomous driving scenarios, including video (5 frames), multi-view images (6 perspectives), and single images, respectively.
% Their autonomous driving scenarios consist of video (5 frames), multi-view images (6 views), and single-image, respectively.
Notably, the answers in the raw data consist of more tokens (e.g., detailed driving suggestions), distinguishing them from unimodal knowledge editing tasks, such as those represented by triple-based formats, which typically feature concise token responses.
% distinguishing them from unimodal knowledge editing tasks, such as those represented by triple-based, which have concise token answers.
This introduces the following challenges: \textbf{1) Suboptimal Editing Performance.} Redundant answers pose difficulties for the implementation of certain editing methods, particularly causal tracing-based approaches such as ROME \citep{rome}. 
\textbf{2) Difficult Evaluation.} Evaluating the accuracy of long text sequences often requires external models to score semantic consistency, which not only suffers from limited accuracy but also incurs high computational costs.
Consequently, simplifying the answers in the raw data becomes a necessary step, as shown in Figure~\ref{fig:overview}.
We prompt Deepseek-v3 \citep{deepseekv3} to condense the original answers.% (The prompt template is shown in Appendix~\ref{appdix:bench}).
Meanwhile, we retain the original answers as meta information for potential use in future research.

%自行补充
\subsubsection{Reliability Data Construction}
% Based on the selected source datasets, we uniformly select a subset as the reliability data.
We observe that the data from DriveLM include predictions of vehicle trajectories, which are not feasible for streamlined and generalized data collection. 
Consequently, we decide to exclude it. 
For CODA-LM, which contains a significant amount of autonomous driving suggestions and image descriptions, we opt to use Deepseek-v3 to self-generate QA pairs.% (The prompt template is shown in Appendix~\ref{appdix:bench}).
We then shuffle all the data and select a subset to serve as reliability data.
Furthermore, to present the metadata from multiple perspectives, we categorize the selected data by scenario types, comprising three main categories and eleven subcategories. 
Specifically, we prompt DeepSeek-v3 to perform classification, which is then verified through rigorous manual inspection, as detailed in Section ~\ref{sec:quality}.

\begin{table*}[t!]
	\centering
	%\scriptsize
	\footnotesize
 %    \small
	% \setlength{\tabcolsep}{4pt}
	%\hfill{}
         % \resizebox{2.0\columnwidth}{!}{
         \scalebox{1.1}{
	\begin{tabular}{l c c c c c c  c c c c c }

		%\toprule
	\toprule
        &  \multirow{2}{*}{\textbf{Method}} &  \multicolumn{2}{c}{Reliability}$\uparrow$  & \multicolumn{2}{c}{T-Generality}$\uparrow$ & \multicolumn{2}{c}{M-Generality}$\uparrow$ & \multicolumn{2}{c}{T-Locality }$\uparrow$ & \multicolumn{2}{c}{M-Locality}$\uparrow$ \\
        \cmidrule{3-4} \cmidrule(lr){5-6} \cmidrule(lr){7-8} \cmidrule(lr){9-10} \cmidrule(lr){11-12}
        & & \multicolumn{1}{c}{Edit.} & \multicolumn{1}{c}{Real.} 
        & \multicolumn{1}{c}{Edit.} & \multicolumn{1}{c}{Real.}
        & \multicolumn{1}{c}{Edit.} & \multicolumn{1}{c}{Real.}
        & \multicolumn{1}{c}{Edit.} & \multicolumn{1}{c}{Real.}
        & \multicolumn{1}{c}{Edit.} & \multicolumn{1}{c}{Real.} \\
        \midrule
		\multicolumn{1}{l}{\scriptsize \textcolor{darkgray}{}} &\multicolumn{11}{c}{\textbf{LLaVA-Onevision}}  \\
		\midrule
        & Prompt & 94.25 & \GG 97.68 & \GG 90.18 & \GG 94.71 & 95.04 & \GG 97.21 & 84.47 & 11.87 & 80.35 & 79.13  \\
          & AdaLora & 78.01 & 56.68 & 72.76 & 54.36 & 75.84 & 53.34 & 85.51 & 40.54 &  81.12 & 69.67 \\
	   & GRACE & \GG 100.00 & 89.52 & 28.91 & 45.83 & 28.16 & 45.45 &\GG 100.00 & \GG 100.00 & \GG 100.00 & \GG 100.00 \\
          & WISE & 99.10 & 75.97 & 86.97 & 70.87 & \GG 95.78 & 73.56 & 94.18 & 56.96 & 99.98 & 82.84\\
		\midrule
		\multicolumn{1}{l}{\scriptsize \textcolor{darkgray}{}} & \multicolumn{11}{c}{\textbf{Qwen2-VL}}  \\
            \midrule
        & Prompt & 90.57 & \GG 97.77 & 84.98 & \GG 94.53 & 90.48 & \GG 97.03 & 89.61 & 61.04 & 72.44 &  66.14  \\
          & AdaLora & 79.89 & 65.03 & 75.68 & 61.69 & 78.76 & 63.64 & 82.27 & 41.37 & 69.37 & 57.33 \\
	   & GRACE & \GG 100.00 & 51.67 & 27.01 & 47.96 & 29.93 & 52.41 & \GG 100.00 & \GG 100.00 & \GG 100.00 & \GG 100.00 \\
          & WISE & 94.18 & 62.98 & \GG 85.20 & 57.14 & \GG 91.99 & 62.15 & 94.23 & 70.13 & 99.85  & 71.33 \\
		\bottomrule

	\end{tabular}
        }
	%\hfill{}
	\caption{Single edit results on the {\ds} under editing evaluation (\textbf{Edit.}) and real-world
evaluation (\textbf{Real.}). \textbf{Reliability} denotes the accuracy of successful editing. \textbf{T-Generality, M-Generality} represents textual and multimodal generality.  \textbf{T-Locality, M-Locality} refer to the textual and multimodal stability.}
 \label{tab:single}
\end{table*}

\subsubsection{Generality Data Construction}
\paragraph{Textual Generality Data.} 
Given the same driving scenario, the text query is modified to test whether the model truly understands the underlying context.
Following prior works, we prompt Deepseek-v3 to rewrite the original query into a semantically similar query while keeping the answer unchanged. %(See the prompt in Appendix~\ref{appdix:bench}).

\paragraph{Multimodal Generality Data.}
% 以往的工作中，要么使用文生图模型生成图像或者人为筛选出相似图片，从而获得视觉泛化的图像数据。而Navigation Cruise领域里使用的都是视频，之前工作中的方法是不切实际的。
% 因此我们提出了假设，假设具有相同问题和回答的数据所对应的视频存在相近的视觉特征。

In previous work, text-to-image generation models were employed to create images, or manually curated similar images were used to obtain visually generalized data. 
However, directly using generative models yields suboptimal performance for the driving domain, which relies exclusively on video data or multiple-view images.
The cost of a fully manual selection of similar autonomous driving videos or images is prohibitively high.
To address this limitation, we hypothesize that videos corresponding to data with identical questions and answers exhibit similar visual characteristics.
Therefore, we only need to match data with identical QA pairs, and then perform non-replacement sampling of two data points per round until exhaustion or only one data point remains. 
Notably, this process will only occur within the same type of visual data.

\subsubsection{Locality Data Construction}
% To evaluate the capability of editing methods in preserving text locality, we select the Natural Questions dataset \citep{nq}, which contains unimodal factual and commonsense knowledge, and randomly sample a subset of it to serve as \textbf{Textual Locality data}. 
% For assessing the preservation of multimodal general capabilities, we select a portion of the VQAv2 dataset \citep{vqa}, which includes general visual question answers, to construct \textbf{Multimodal Locality data}.
To evaluate the ability of editing methods to preserve textual locality, we utilize the Natural Questions dataset \citep{nq}, which encompasses unimodal factual and commonsense knowledge. A random subset of this dataset is sampled to constitute the Textual Locality data.
For assessing the preservation of general multimodal capabilities, we select a portion of the VQAv2 dataset \citep{vqa}, comprising general visual question-answer pairs, to form the Multimodal Locality data.

\subsection{Quality Control}
\label{sec:quality}
% 每一部分都要详细检查
% 由于上述的过程见图，都是自动化生成，人工保证数据的质量是必要的。人工审核
% 1.对简写的答案 进行校准
% 2.对coda-lm生成的 qa进行校准
% 3.对t-gen 校准
% 4.对m-gen生成的图片进行过滤
% 5.对分类进行人工打标+多名标注员
As outlined in the aforementioned process and illustrated in Figure~\ref{fig:overview}, the entire pipeline is fully automated. However, ensuring data quality through manual verification remains a critical step. 
To this end, we conduct manual verification and calibration for all processes involving AI models, including answer simplification, QA pairs generation from CODA-LM, textual generality data generation, and scenario categorization. 
Furthermore, for multimodal generality data, we manually sample 20\% of the data to assess similarity and filter out those instances with no resemblance.
To ensure the consistency between the rationales in the generality data answers and the reliability data, we filter out mismatched samples by referring to the original answers.
The data quality control tasks are evenly distributed among three annotators with graduate-level education, who independently performed the calibration. 
In cases of uncertainty, the annotators first independently judge on whether modifications are necessary, and then resolve any disagreements through a majority vote. 
The inter-annotator consistency, measured by Fleiss’s Kappa (\(\kappa\)), demonstrates substantial agreement (\(\kappa\) = 0.802), based on a randomly selected sample of 200 annotated data points, indicating a high level of consistency within the range of 0.80 ≤ \(\kappa\) ≤ 1.00.

% To this end, we manually filter out video data that is unclear or text generalization data with ambiguous expressions.
% The dataset was evenly distributed among three graduate-level annotators, each of whom independently categorized the data. In cases where uncertainty arose, initial decisions were made by individual annotators and subsequently resolved through majority voting. The inter-annotator reliability, measured using Fleiss's Kappa (κ), demonstrated substantial agreement with a score of κ=0.822. This value falls within the range of 0.80≤κ≤1.00, indicating a high level of concordance among annotators and underscoring the robustness of our quality control measures.

\section{Evaluation}

\subsection{Experimental settings}
We conduct experiments on the most advanced LMMs to date, which are LLaVA-OneVision \cite{llavaonevision} and Qwen2-VL \cite{qwen2vl}.
Both models utilize Qwen2-7B as their LLM component.
Meanwhile, four classical knowledge editing baselines, including Prompt, AdaLora, GRACE, and WISE are employed to update the knowledge of LMMs. 
%Further details can be found in the Appendix~\ref{appdix:baseline}.

Our analysis covers both the \textbf{Single Editing} and \textbf{Lifelong Editing} settings to thoroughly assess the performance of different baselines in \ds.
Single editing involves updating and evaluating the model immediately after receiving each individual driving data instance. 
In contrast, lifelong editing represents a more realistic driving scenario, where multiple driving data instances are collected during the vehicle's moving, requiring continuous integration of this knowledge into the model followed by evaluation. 
We reasonably assume that the visual processing component (e.g., Vision Transformer) of LMMs provides reliable visual information. Therefore, we focus only on editing the LLM component to modify its understanding of visual input.

To comprehensively evaluate the effectiveness of editing methods on ADS, we adopt two evaluation metrics: 
(1) Editing evaluation (\textbf{Edit}), where tokens are generated in a teacher-forcing manner and compared at the token level \citep{rome}.
(2) Real-world evaluation (\textbf{Real}), where the edited model performs free-form generation, and a LLM is prompted to assess the semantic consistency between the generated outputs and the ground-truth answers \citep{realworldedit}.
Due to cost and resource constraints, we report the Real metric only for the main single-edit experiment, while the remaining experiments are evaluated using the Edit metric by default.

\subsection{Main Results}
% 按两个指标的结果来讲
We present the results for the single-edit scenario below; the results for lifelong editing can be found on Github\footnote{\url{https://github.com/zjunlp/EasyEdit/blob/main/examples/ADSEdit.md}}.

\paragraph{Results of Editing evaluation.}

Memory-based editing methods, such as GRACE and WISE, have proven highly effective in modulating the behavior of LMMs for autonomous driving task prediction. 
Notably, GRACE achieves a 100\% modification rate in both models, yet performs poorly in terms of generality.
WISE demonstrates a well-balanced performance across reliability, generality, and locality.
Although the ground truth answer is directly provided in the input text, the reliability and generality of the Prompt remains suboptimal. 
Meanwhile, Prompt performs poorly on the locality, which can be attributed to its sensitivity to the ground-truth answers in the input.
Under a limited resources, AdaLora struggles to achieve satisfactory reliability and generality through model parameter updates.
Moreover, parameter modifications significantly impact Adalora's performance on the locality metric.

% Evaluating the generalization of editing methods to other autonomous driving scenarios and queries is essential.
% However, GRACE exhibits the worst prediction performance on Generality, with accuracy dropping below 30\% in both LMMs, compared to its Reliability.
% This is attributed to the difficulty of GRACE's codebook in capturing the representational differences of long-sequence multimodal inputs.
% The remaining three baselines show similar Generality and Reliability results. 
% Among them, WISE and Prompt achieve approximately 90\% on Generality, demonstrating their stable generalization capabilities.
% AdaLora still struggle to generalize the other driving scenarios.
% Furthermore, M-Generality of the baselines exceeds T-Generality, suggesting that LMMs tend to focus more on text tokens while making less use of visual tokens.
% This observation may inspire further research on enhancing the efficient utilization of redundant visual representations in LMMs.

% While preserving the model's local capabilities, GRACE demonstrates a significant advantage in Locality.
% WISE, which employs a dual-memory mechanism, also maintains good Locality, achieving nearly 100\%. 
% In contrast, AdaLora and Prompt somewhat disrupt the original model's Locality. 
% Additionally, we observe that both AdaLora and Prompt exhibit lower M-Locality than T-Locality, which is due to the changes in the LMM's behavior caused by the multimodal inputs in autonomous driving scenarios.

\begin{figure}[hb!] %H为当前位置，!htb为忽略美学标准，htbp为浮动图形
\centering %图片居中
\includegraphics[width=0.48\textwidth]{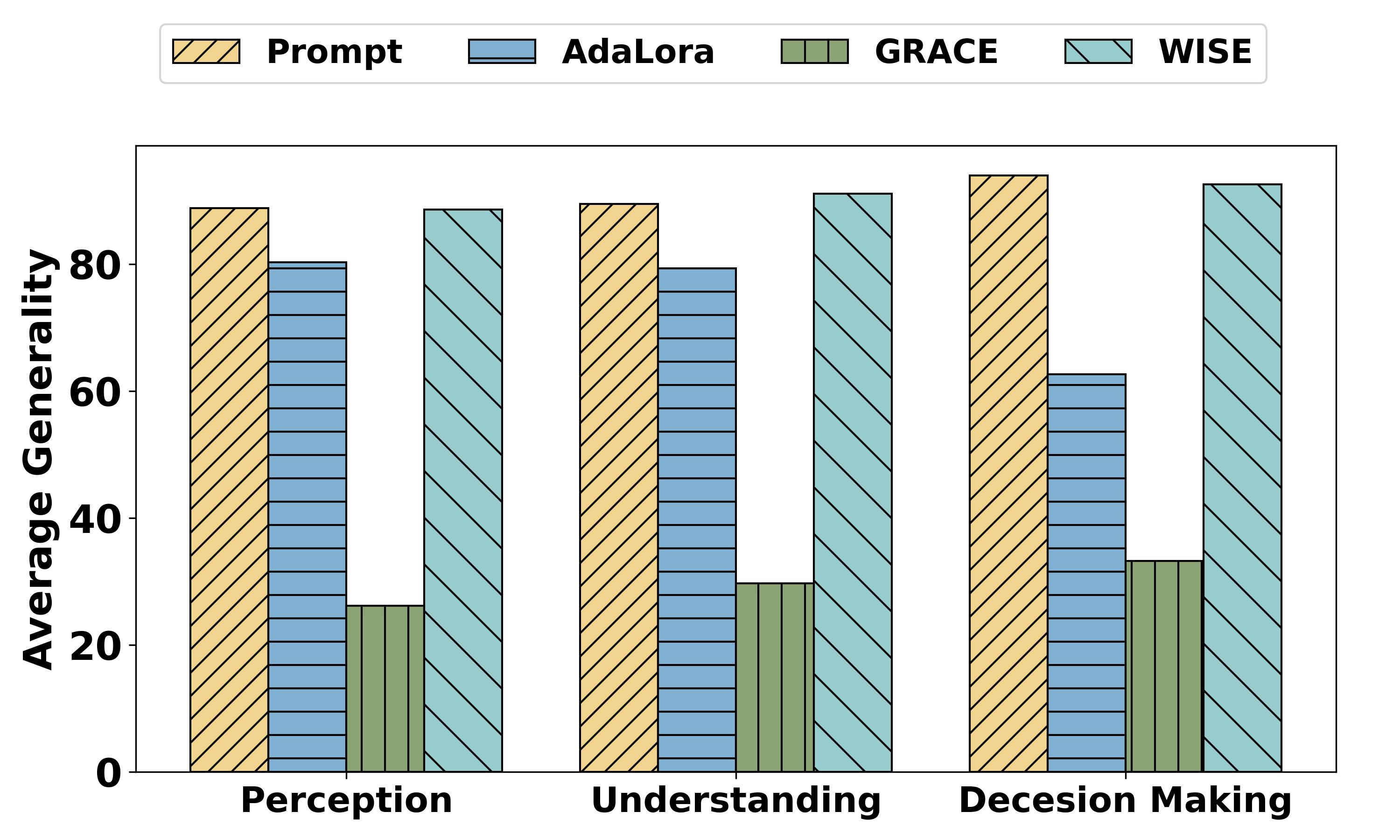} %
% 插入图片，[]中设置图片大小，{}中是图片文件名
\caption{
The average generality metric of single editing across different scenarios.}
\label{fig:task_types}
\vspace{-0.5cm}
\end{figure}

\paragraph{Results of Real-world evaluation.}
Although Prompt underperforms GRACE and WISE on conventional editing metrics, it achieves the best performance in terms of reliability and generalization under real-world evaluation settings.
This is primarily because the model does not strictly replicate the ground-truth answers but instead generates outputs that are semantically aligned with them. 
In contrast, the other three editing methods exhibit a noticeable decline in performance when transitioning from editing metrics to real-world evaluation, indicating \textbf{limitations in their reliability and generalization for ADS applications.}
In terms of locality metrics, GRACE and WISE demonstrate strong performance, whereas Prompt underperforms.

\subsection{Analysis}

\paragraph{LLaVA-OneVision vs. Qwen2-VL: Which is Easier to Edit?} Although both models use Qwen2 as Large Language Model, LLaVA-Onevision achieve better results across most metrics in terms of reliability and generality.
Specifically, on Prompt and WISE methods, Qwen2-VL demonstrates stronger performance in retaining the original predicted answers.
Qwen2-VL slightly outperforms LLaVA-OneVision in terms of reliability and generality by using AdaLora. However, locality is significantly compromised compared to LLaVA-OneVision.
Furthermore, in real-world evaluation scores, we observe that Qwen2-VL tends to retain its original outputs and exhibits resistance to editing.

\begin{figure}[htb!] %H为当前位置，!htb为忽略美学标准，htbp为浮动图形
\centering %图片居中
\includegraphics[width=0.48\textwidth]{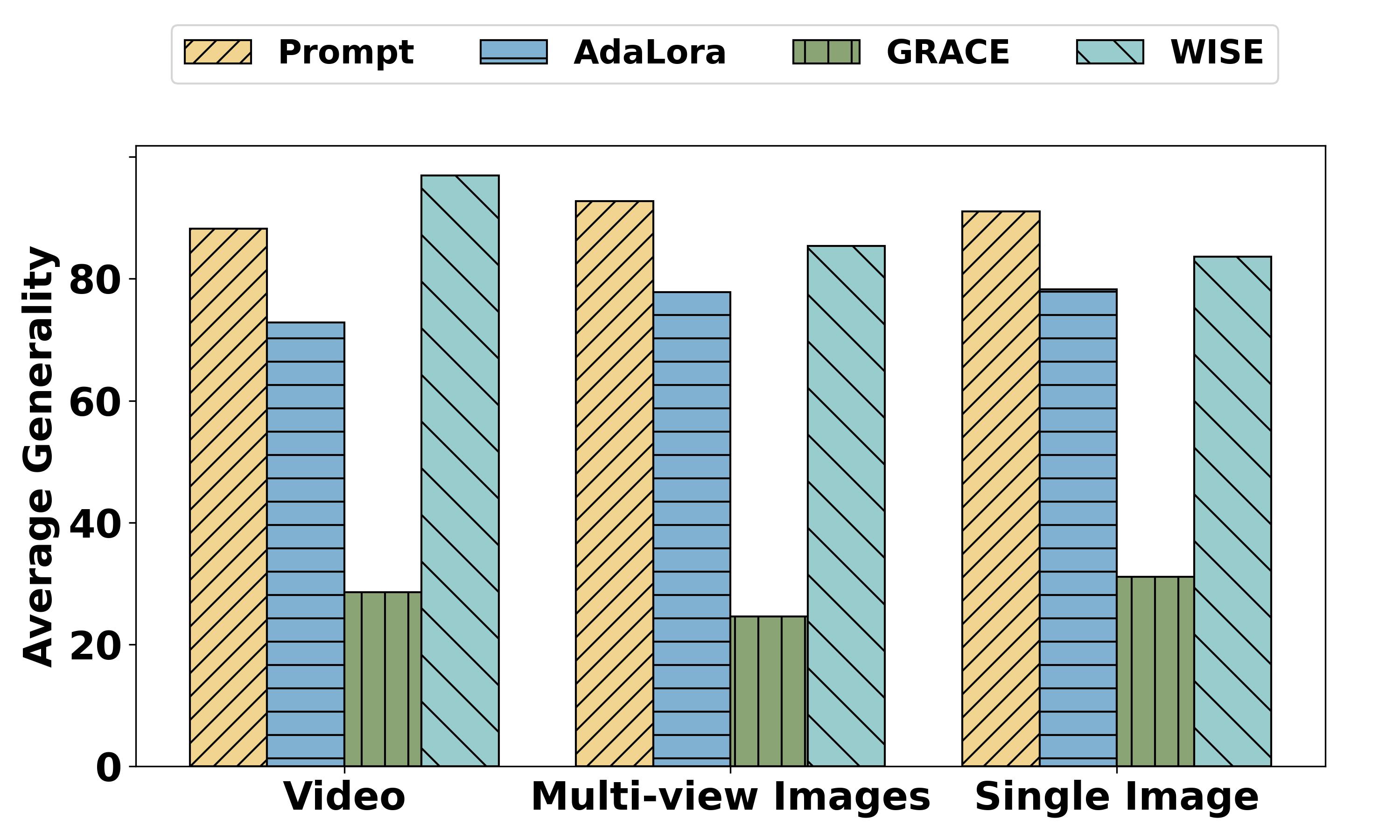} %
% 插入图片，[]中设置图片大小，{}中是图片文件名
\caption{
The average generality metric of single editing across different data types.}
\label{fig:data_types}
\vspace{-0.5cm}
\end{figure}

\begin{figure*}[ht!] 
\centering
\includegraphics[width=0.86\textwidth]{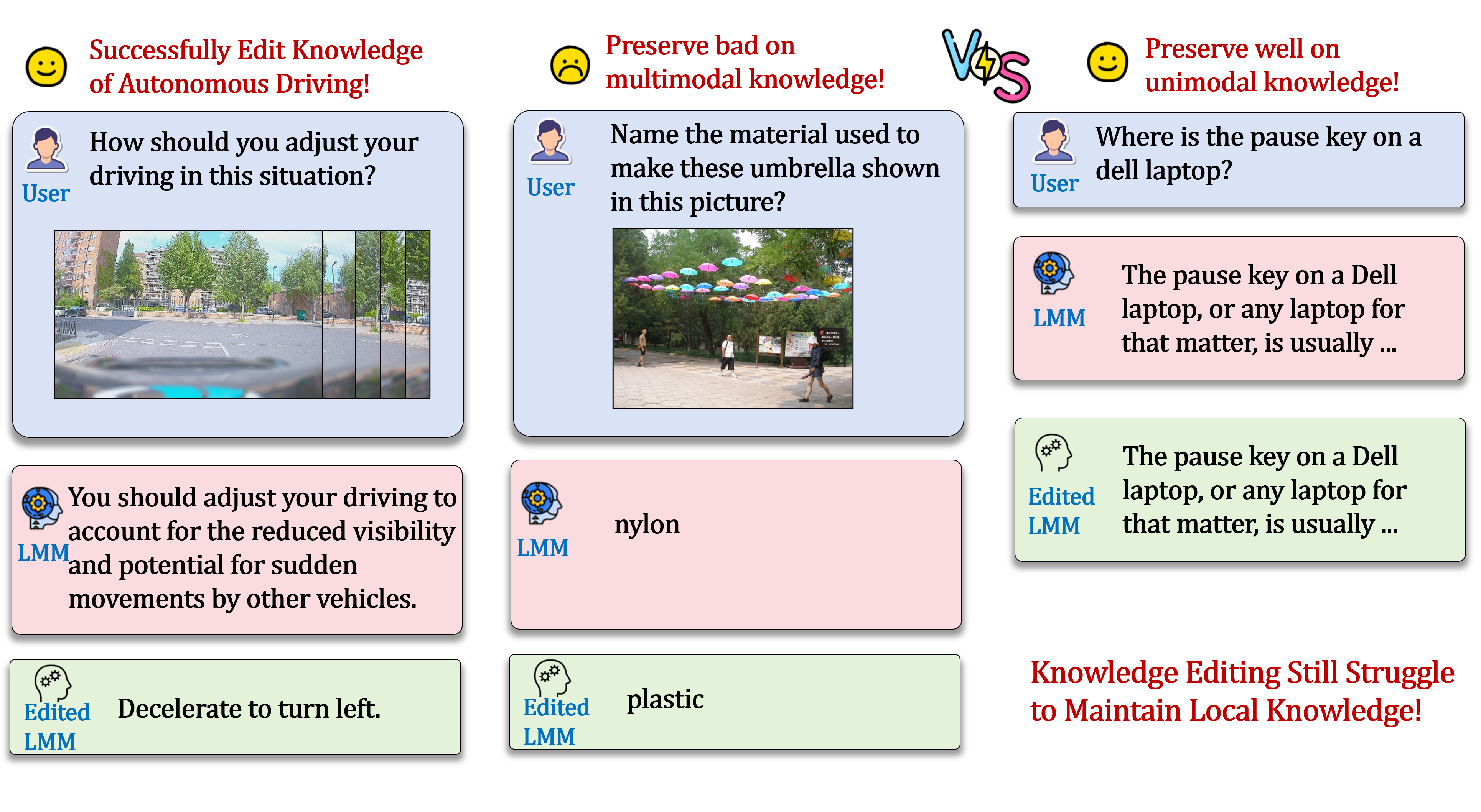}
\caption{Cases analysis of editing LLaVA-OneVision with WISE.}
\label{fig:case} 
\end{figure*}

% 场景分析
\paragraph{Which Scenario Type Knowledges are Easier to Update?}
% 柱状图
% 基于参数更新的 决策最难更新
% 感知和理解的效果都差不多
% 基于memory 的reliability的效果都差不多
We choose the average generality metric, which is the mean of the text and multimodal generality of Qwen2-VL and LLaVA-OneVision, to assess the performance of the edits.
(Since both WISE and GRACE achieve Reliability close to 100\%, analyzing the Reliability metric becomes less meaningful. This observation applies to all subsequent analyses in this study.)
According to the averaged metrics across three scenarios illustrated in Figure~\ref{fig:task_types}, for AdaLora, decision scenario is more challenging to learn compared to perception and understanding scenario, given a fixed number of training epochs. 
This can be attributed to the higher complexity of decision making data. 
In contrast, Prompt, GRACE, and WISE show relatively consistent performance metrics across all scenarios, with minimal variation in their generality.
\textbf{Overall, these knowledge editing methods are broadly applicable to various driving scenarios, effectively updating knowledge in LMMs.}

\begin{table}[h]
\centering
\small
% \resizebox{1.0\columnwidth}{!}{
% \resizebox{1.0\textwidth}{!}{% <------ Don't forget this %
\begin{tabular}{ccc}
\toprule
& \bf Low Video Frame rate & \bf Max Video Frame rate  \\ 
\midrule
Prompt & 90.14 & \textbf{92.61} \\
AdaLora & 70.04 & \textbf{74.30} \\
GRACE & \textbf{34.33} & 28.54 \\
WISE & \textbf{99.49} & 91.37 \\
\bottomrule
\end{tabular}
% }
% }
\caption{
The average generality results from different video frames.
}
\label{tab:reduce}
\end{table}

\paragraph{Which Data Type is Easier to Edit?}
We compute the baseline performance across different types of modality, including video, multi-view images and single image, as shown in Figure~\ref{fig:data_types}. 
WISE demonstrates the highest performance on video data, likely due to its memory mechanism to store and update knowledge in temporal changes in driving scenario.
Both Prompt and AdaLoRA exhibit a gradual increase in generality as the length of the visual sequence decreases.
% We compute the baseline performance across different types of modality, including video, multi-view images and single image, as shown in Figure~\ref{fig:data_types}. 
% WISE demonstrates the highest performance on video data, likely due to its memory mechanism to store and update knowledge in temporal changes in driving scenario.
% Both Prompt and AdaLoRA exhibit a gradual increase in generality as the length of the visual sequence decreases.

% AdaLora performs sub-optimally on the more challenging video data. On the other hand, memory-based methods exhibit poor performance on single-image type data. We hypothesize that the longer visual tokens in video data make it difficult for AdaLora to effectively capture the underlying pattern features. In contrast, memory-based methods excel at storing updates to feature flows.

\paragraph{Does Reducing Video Frames Impact the Effectiveness of Knowledge Editing?}
% 折线图
To assess the impact of video frames on editing performance, we sequentially test videos with 1 to 5 frames as input in single editing setting. 
The average generality on video data with 1-4 frames is denoted as the effect of the knowledge editing method under low video frame rate conditions, whereas video data with 5 frames is referred to as the maximum frame rate condition.
Fewer video frames correspond to fewer visual tokens, enabling LMMs to process users' requests more quickly. 
Although this reduction in frames results in some loss of information, WISE demonstrates even better performance.
Furthermore, the results of both Prompt and AdaLora are not significantly affected under conditions of both low and maximum frame rate.
This suggests that \textbf{knowledge editing methods in the autonomous driving domain, particularly for video data, can effectively balance processing speed and performance.}

% Fewer video frames correspond to fewer visual tokens, leading to more missing contextual information and poorer generalization of the editing process, as shown in Table~\ref{tab:reduce}.
% However, the reliability remains nearly unchanged across these variations.

% bad case分析
\paragraph{Cases Analysis of Knowledge Editing.}
From Figure~\ref{fig:case}, we observe that the edited LMM tends to maintain better locality in the unimodal domain compared to the multimodal one. 
Interestingly, despite only editing the language model component of the LMM, the model's visual understanding is impaired. 
This underscores the distinct differences between the LMM's capabilities in visual understanding and text-based knowledge storage.
% As shown in Figure~\ref{fig:case}, we observe that the edited LMM tends to maintain better locality in the unimodal domain compared to the multimodal one. 
% Notably, although only the language model component of the LMM was modified, a noticeable decline in the model's visual understanding was observed. This highlights a fundamental distinction between the LMM's visual comprehension capabilities and its text-based knowledge storage, underscoring the separate and potentially independent functionalities of these two domains.

\section{Related Work}
% 前面多写一点 related work需要精简

\subsection{Knowledge Editing}
Recent advances in knowledge editing have emerged as a pivotal research direction in knowledge updating \citep{2024Large,DBLP:journals/corr/abs-2407-06249,DBLP:journals/corr/abs-2401-10471,DBLP:conf/emnlp/WuPWL24,DBLP:conf/cikm/XuZZLL00WY0C024}.
% DBLP:conf/acl/LiC24,DBLP:journals/corr/abs-2406-17764,DBLP:conf/coling/WeiDPDSC25,DBLP:conf/coling/ZhangLMZ0X025,DBLP:conf/coling/AliDWQ025,hwang2025personality,zhao2025fleke,markowitz2025k,yang2025mirage,dong2025memit,pan2025precise,wang2025revealing,liuunlocking,zhang2024uncovering,xu2025constraining,bi2025parameters,he2025knowledge,li2025mindbridge,shen2025llm,realworldedit}.
Following \citet{yao2023editing}, we systematize current methodologies into two principal paradigms:
(1) Parameter Modification Approaches:
These methods directly alter the model's internal representations through targeted interventions \citep{rome,memit,alphaedit,o-edit,mend,instructedit,lora, adalora, qlora}. % malmen feng2025geoedit
(2) Parameter Preservation Approaches:
These strategies maintain original parameters while updating new knowledge \citep{ike,wise,grace}. % li2025mindbridge,lte
The emergence of Large Multimodal Models (LMMs) has also advanced the field of multimodal knowledge editing.
These works mainly focus on the editing of multimodal common knowledge \citep{cheng2023mmedit} or multimodal factual knowledge \citep{mike,vlkeb}. % pan2024towards,wang2024can mcmke comprehendedit 
However, the application of multimodal knowledge editing in the context of domain-specific knowledge, like autonomous driving remains underexplored.

\subsection{Large Multi-modal Models for Autonomous Driving}
% ada-survey
The recent rise of large multimodal models (LMMs) in autonomous driving~\citep{NuScenes-QA} has spurred the development of advanced frameworks. 
CODA-LM~\citep{CODA-LM} and DriveLM~\citep{drivelm} integrate hierarchical reasoning and graph-structured visual QA frameworks.
Meanwhile, LingoQA~\citep{lingoqa} expands free-form QA capabilities by integrating action justification and scene understanding.
However, these models still face challenges in real-world deployment due to static knowledge representations and modality imbalance.

\section{Conclusion}
% In this paper, we present to leverage knowledge editing techniques to address the challenges faced by Large Multimodal Models in autonomous driving scenarios. 
% ADS-Edit, a new benchmark specifically is designed for evaluating knowledge editing methods in this domain.
% Through extensive experimental analysis, we systematically compare and evaluate the effectiveness of various editing methods, while providing in-depth insights into the underlying reasons for the observed performance variations and failure cases.
% Moreover, our findings highlight that developing editing methods that are both well-balanced and high-performing remains a significant challenge in the autonomous driving systems (ADS).
In this paper, we propose leveraging knowledge editing techniques to address the challenges faced by Large Multimodal Models in autonomous driving scenarios. To this end, we introduce ADS-Edit, a novel benchmark specifically designed to evaluate knowledge editing methods within this domain. Through extensive experimental analysis, we systematically compare and assess the effectiveness of various editing approaches, providing comprehensive insights into the underlying causes of performance variations and failure cases. 
Moreover, our findings highlight that developing editing methods that are both well-balanced and high-performing remains a significant challenge in the autonomous driving systems.

\begin{acks}
We would like to express our sincere gratitude to the anonymous reviewers for their thoughtful and constructive feedback. This work was supported by the National Natural Science Foundation of China (No. 62206246, No. NSFCU23B2055, No. NSFCU19B2027), the Fundamental Research Funds for the Central Universities (226-2023- 00138), Ningbo Natural Science Foundation (2024J020), Yongjiang Talent Introduction Programme (2021A-156-G), and Information Technology Center and State Key Lab of CAD\&CG, Zhejiang University. This work was supported by Ant Group and Zhejiang University - Ant Group Joint Laboratory of Knowledge Graph.
\end{acks}
\bibliographystyle{ACM-Reference-Format}
\balance
\bibliography{custom}


\begin{thebibliography}{45}

%%% the Web address.

\ifx \showCODEN    \undefined \def \showCODEN     #1{\unskip}     \fi
\ifx \showISBNx    \undefined \def \showISBNx     #1{\unskip}     \fi
\ifx \showISBNxiii \undefined \def \showISBNxiii  #1{\unskip}     \fi
\ifx \showISSN     \undefined \def \showISSN      #1{\unskip}     \fi
\ifx \showLCCN     \undefined \def \showLCCN      #1{\unskip}     \fi
\ifx \shownote     \undefined \def \shownote      #1{#1}          \fi
\ifx \showarticletitle \undefined \def \showarticletitle #1{#1}   \fi
\ifx \showURL      \undefined \def \showURL       {\relax}        \fi
% invisible to TeX
\providecommand\bibfield[2]{#2}
\providecommand\bibinfo[2]{#2}
\providecommand\natexlab[1]{#1}
\providecommand\showeprint[2][]{arXiv:#2}

\bibitem[Antol et~al\mbox{.}(2015)]%
        {vqa}
\bibfield{author}{\bibinfo{person}{Stanislaw Antol}, \bibinfo{person}{Aishwarya Agrawal}, \bibinfo{person}{Jiasen Lu}, \bibinfo{person}{Margaret Mitchell}, \bibinfo{person}{Dhruv Batra}, \bibinfo{person}{C.~Lawrence Zitnick}, {and} \bibinfo{person}{Devi Parikh}.} \bibinfo{year}{2015}\natexlab{}.
\newblock \showarticletitle{VQA: Visual Question Answering}. In \bibinfo{booktitle}{\emph{International Conference on Computer Vision (ICCV)}}.
\newblock


\bibitem[Cai and Cao(2024)]%
        {o-edit}
\bibfield{author}{\bibinfo{person}{Yuchen Cai} {and} \bibinfo{person}{Ding Cao}.} \bibinfo{year}{2024}\natexlab{}.
\newblock \showarticletitle{O-Edit: Orthogonal Subspace Editing for Language Model Sequential Editing}.
\newblock \bibinfo{journal}{\emph{CoRR}}  \bibinfo{volume}{abs/2410.11469} (\bibinfo{year}{2024}).
\newblock
\href{https://doi.org/10.48550/ARXIV.2410.11469}{doi:\nolinkurl{10.48550/ARXIV.2410.11469}}
\showeprint[arXiv]{2410.11469}


\bibitem[Chen(2024)]%
        {2024Large}
\bibfield{author}{\bibinfo{person}{Huajun Chen}.} \bibinfo{year}{2024}\natexlab{}.
\newblock \showarticletitle{Large Knowledge Model: Perspectives and Challenges}.
\newblock \bibinfo{journal}{\emph{Data Intelligence}} \bibinfo{number}{3} (\bibinfo{year}{2024}).
\newblock


\bibitem[Cheng et~al\mbox{.}(2023)]%
        {cheng2023mmedit}
\bibfield{author}{\bibinfo{person}{Siyuan Cheng}, \bibinfo{person}{Bozhong Tian}, \bibinfo{person}{Qingbin Liu}, \bibinfo{person}{Xi Chen}, \bibinfo{person}{Yongheng Wang}, \bibinfo{person}{Huajun Chen}, {and} \bibinfo{person}{Ningyu Zhang}.} \bibinfo{year}{2023}\natexlab{}.
\newblock \showarticletitle{Can We Edit Multimodal Large Language Models?}. In \bibinfo{booktitle}{\emph{Proceedings of the 2023 Conference on Empirical Methods in Natural Language Processing, {EMNLP} 2023, Singapore, December 6-10, 2023}}, \bibfield{editor}{\bibinfo{person}{Houda Bouamor}, \bibinfo{person}{Juan Pino}, {and} \bibinfo{person}{Kalika Bali}} (Eds.). \bibinfo{publisher}{Association for Computational Linguistics}, \bibinfo{pages}{13877--13888}.
\newblock
\href{https://doi.org/10.18653/V1/2023.EMNLP-MAIN.856}{doi:\nolinkurl{10.18653/V1/2023.EMNLP-MAIN.856}}


\bibitem[Dettmers et~al\mbox{.}(2023)]%
        {qlora}
\bibfield{author}{\bibinfo{person}{Tim Dettmers}, \bibinfo{person}{Artidoro Pagnoni}, \bibinfo{person}{Ari Holtzman}, {and} \bibinfo{person}{Luke Zettlemoyer}.} \bibinfo{year}{2023}\natexlab{}.
\newblock \showarticletitle{QLoRA: Efficient Finetuning of Quantized LLMs}. In \bibinfo{booktitle}{\emph{Advances in Neural Information Processing Systems 36: Annual Conference on Neural Information Processing Systems 2023, NeurIPS 2023, New Orleans, LA, USA, December 10 - 16, 2023}}, \bibfield{editor}{\bibinfo{person}{Alice Oh}, \bibinfo{person}{Tristan Naumann}, \bibinfo{person}{Amir Globerson}, \bibinfo{person}{Kate Saenko}, \bibinfo{person}{Moritz Hardt}, {and} \bibinfo{person}{Sergey Levine}} (Eds.).
\newblock
\urldef\tempurl%
\url{http://papers.nips.cc/paper\_files/paper/2023/hash/1feb87871436031bdc0f2beaa62a049b-Abstract-Conference.html}
\showURL{%
\tempurl}


\bibitem[et~al(2024)]%
        {deepseekv3}
\bibfield{author}{\bibinfo{person}{DeepSeek-AI et al}.} \bibinfo{year}{2024}\natexlab{}.
\newblock \bibinfo{title}{DeepSeek-V3 Technical Report}.
\newblock
\showeprint[arxiv]{2412.19437}~[cs.CL]
\urldef\tempurl%
\url{https://arxiv.org/abs/2412.19437}
\showURL{%
\tempurl}


\bibitem[Fang et~al\mbox{.}(2024a)]%
        {DBLP:journals/corr/abs-2410-02355}
\bibfield{author}{\bibinfo{person}{Junfeng Fang}, \bibinfo{person}{Houcheng Jiang}, \bibinfo{person}{Kun Wang}, \bibinfo{person}{Yunshan Ma}, \bibinfo{person}{Xiang Wang}, \bibinfo{person}{Xiangnan He}, {and} \bibinfo{person}{Tat{-}Seng Chua}.} \bibinfo{year}{2024}\natexlab{a}.
\newblock \showarticletitle{AlphaEdit: Null-Space Constrained Knowledge Editing for Language Models}.
\newblock \bibinfo{journal}{\emph{CoRR}}  \bibinfo{volume}{abs/2410.02355} (\bibinfo{year}{2024}).
\newblock
\href{https://doi.org/10.48550/ARXIV.2410.02355}{doi:\nolinkurl{10.48550/ARXIV.2410.02355}}
\showeprint[arXiv]{2410.02355}


\bibitem[Fang et~al\mbox{.}(2024b)]%
        {alphaedit}
\bibfield{author}{\bibinfo{person}{Junfeng Fang}, \bibinfo{person}{Houcheng Jiang}, \bibinfo{person}{Kun Wang}, \bibinfo{person}{Yunshan Ma}, \bibinfo{person}{Xiang Wang}, \bibinfo{person}{Xiangnan He}, {and} \bibinfo{person}{Tat{-}Seng Chua}.} \bibinfo{year}{2024}\natexlab{b}.
\newblock \showarticletitle{AlphaEdit: Null-Space Constrained Knowledge Editing for Language Models}.
\newblock \bibinfo{journal}{\emph{CoRR}}  \bibinfo{volume}{abs/2410.02355} (\bibinfo{year}{2024}).
\newblock
\href{https://doi.org/10.48550/ARXIV.2410.02355}{doi:\nolinkurl{10.48550/ARXIV.2410.02355}}
\showeprint[arXiv]{2410.02355}


\bibitem[Gupta et~al\mbox{.}(2024)]%
        {gupta2024model}
\bibfield{author}{\bibinfo{person}{Akshat Gupta}, \bibinfo{person}{Anurag Rao}, {and} \bibinfo{person}{Gopala Anumanchipalli}.} \bibinfo{year}{2024}\natexlab{}.
\newblock \showarticletitle{Model editing at scale leads to gradual and catastrophic forgetting}.
\newblock \bibinfo{journal}{\emph{arXiv preprint arXiv:2401.07453}} (\bibinfo{year}{2024}).
\newblock


\bibitem[Hartvigsen et~al\mbox{.}(2023)]%
        {grace}
\bibfield{author}{\bibinfo{person}{Tom Hartvigsen}, \bibinfo{person}{Swami Sankaranarayanan}, \bibinfo{person}{Hamid Palangi}, \bibinfo{person}{Yoon Kim}, {and} \bibinfo{person}{Marzyeh Ghassemi}.} \bibinfo{year}{2023}\natexlab{}.
\newblock \showarticletitle{Aging with {GRACE:} Lifelong Model Editing with Discrete Key-Value Adaptors}. In \bibinfo{booktitle}{\emph{Advances in Neural Information Processing Systems 36: Annual Conference on Neural Information Processing Systems 2023, NeurIPS 2023, New Orleans, LA, USA, December 10 - 16, 2023}}, \bibfield{editor}{\bibinfo{person}{Alice Oh}, \bibinfo{person}{Tristan Naumann}, \bibinfo{person}{Amir Globerson}, \bibinfo{person}{Kate Saenko}, \bibinfo{person}{Moritz Hardt}, {and} \bibinfo{person}{Sergey Levine}} (Eds.).
\newblock
\urldef\tempurl%
\url{http://papers.nips.cc/paper\_files/paper/2023/hash/95b6e2ff961580e03c0a662a63a71812-Abstract-Conference.html}
\showURL{%
\tempurl}


\bibitem[Hu et~al\mbox{.}(2022)]%
        {lora}
\bibfield{author}{\bibinfo{person}{Edward~J. Hu}, \bibinfo{person}{Yelong Shen}, \bibinfo{person}{Phillip Wallis}, \bibinfo{person}{Zeyuan Allen{-}Zhu}, \bibinfo{person}{Yuanzhi Li}, \bibinfo{person}{Shean Wang}, \bibinfo{person}{Lu Wang}, {and} \bibinfo{person}{Weizhu Chen}.} \bibinfo{year}{2022}\natexlab{}.
\newblock \showarticletitle{LoRA: Low-Rank Adaptation of Large Language Models}. In \bibinfo{booktitle}{\emph{The Tenth International Conference on Learning Representations, {ICLR} 2022, Virtual Event, April 25-29, 2022}}. \bibinfo{publisher}{OpenReview.net}.
\newblock
\urldef\tempurl%
\url{https://openreview.net/forum?id=nZeVKeeFYf9}
\showURL{%
\tempurl}


\bibitem[Huang et~al\mbox{.}(2024b)]%
        {vlkeb}
\bibfield{author}{\bibinfo{person}{Han Huang}, \bibinfo{person}{Haitian Zhong}, \bibinfo{person}{Tao Yu}, \bibinfo{person}{Qiang Liu}, \bibinfo{person}{Shu Wu}, \bibinfo{person}{Liang Wang}, {and} \bibinfo{person}{Tieniu Tan}.} \bibinfo{year}{2024}\natexlab{b}.
\newblock \showarticletitle{{VLKEB:} {A} Large Vision-Language Model Knowledge Editing Benchmark}. In \bibinfo{booktitle}{\emph{Advances in Neural Information Processing Systems 38: Annual Conference on Neural Information Processing Systems 2024, NeurIPS 2024, Vancouver, BC, Canada, December 10 - 15, 2024}}, \bibfield{editor}{\bibinfo{person}{Amir Globersons}, \bibinfo{person}{Lester Mackey}, \bibinfo{person}{Danielle Belgrave}, \bibinfo{person}{Angela Fan}, \bibinfo{person}{Ulrich Paquet}, \bibinfo{person}{Jakub~M. Tomczak}, {and} \bibinfo{person}{Cheng Zhang}} (Eds.).
\newblock
\urldef\tempurl%
\url{http://papers.nips.cc/paper\_files/paper/2024/hash/1198b53fa686831d5f0c0860d7ec4f34-Abstract-Datasets\_and\_Benchmarks\_Track.html}
\showURL{%
\tempurl}


\bibitem[Huang et~al\mbox{.}(2024a)]%
        {drivemm}
\bibfield{author}{\bibinfo{person}{Zhijian Huang}, \bibinfo{person}{Chengjian Feng}, \bibinfo{person}{Feng Yan}, \bibinfo{person}{Baihui Xiao}, \bibinfo{person}{Zequn Jie}, \bibinfo{person}{Yujie Zhong}, \bibinfo{person}{Xiaodan Liang}, {and} \bibinfo{person}{Lin Ma}.} \bibinfo{year}{2024}\natexlab{a}.
\newblock \showarticletitle{DriveMM: All-in-One Large Multimodal Model for Autonomous Driving}.
\newblock \bibinfo{journal}{\emph{CoRR}}  \bibinfo{volume}{abs/2412.07689} (\bibinfo{year}{2024}).
\newblock
\href{https://doi.org/10.48550/ARXIV.2412.07689}{doi:\nolinkurl{10.48550/ARXIV.2412.07689}}
\showeprint[arXiv]{2412.07689}


\bibitem[Jiang et~al\mbox{.}(2025)]%
        {jiang2025anyedit}
\bibfield{author}{\bibinfo{person}{Houcheng Jiang}, \bibinfo{person}{Junfeng Fang}, \bibinfo{person}{Ningyu Zhang}, \bibinfo{person}{Guojun Ma}, \bibinfo{person}{Mingyang Wan}, \bibinfo{person}{Xiang Wang}, \bibinfo{person}{Xiangnan He}, {and} \bibinfo{person}{Tat-seng Chua}.} \bibinfo{year}{2025}\natexlab{}.
\newblock \showarticletitle{AnyEdit: Edit Any Knowledge Encoded in Language Models}.
\newblock \bibinfo{journal}{\emph{arXiv preprint arXiv:2502.05628}} (\bibinfo{year}{2025}).
\newblock


\bibitem[Kwiatkowski et~al\mbox{.}(2019)]%
        {nq}
\bibfield{author}{\bibinfo{person}{Tom Kwiatkowski}, \bibinfo{person}{Jennimaria Palomaki}, \bibinfo{person}{Olivia Redfield}, \bibinfo{person}{Michael Collins}, \bibinfo{person}{Ankur~P. Parikh}, \bibinfo{person}{Chris Alberti}, \bibinfo{person}{Danielle Epstein}, \bibinfo{person}{Illia Polosukhin}, \bibinfo{person}{Jacob Devlin}, \bibinfo{person}{Kenton Lee}, \bibinfo{person}{Kristina Toutanova}, \bibinfo{person}{Llion Jones}, \bibinfo{person}{Matthew Kelcey}, \bibinfo{person}{Ming{-}Wei Chang}, \bibinfo{person}{Andrew~M. Dai}, \bibinfo{person}{Jakob Uszkoreit}, \bibinfo{person}{Quoc Le}, {and} \bibinfo{person}{Slav Petrov}.} \bibinfo{year}{2019}\natexlab{}.
\newblock \showarticletitle{Natural Questions: a Benchmark for Question Answering Research}.
\newblock \bibinfo{journal}{\emph{Trans. Assoc. Comput. Linguistics}}  \bibinfo{volume}{7} (\bibinfo{year}{2019}), \bibinfo{pages}{452--466}.
\newblock
\href{https://doi.org/10.1162/TACL\_A\_00276}{doi:\nolinkurl{10.1162/TACL\_A\_00276}}


\bibitem[Li et~al\mbox{.}(2024d)]%
        {llavaonevision}
\bibfield{author}{\bibinfo{person}{Bo Li}, \bibinfo{person}{Yuanhan Zhang}, \bibinfo{person}{Dong Guo}, \bibinfo{person}{Renrui Zhang}, \bibinfo{person}{Feng Li}, \bibinfo{person}{Hao Zhang}, \bibinfo{person}{Kaichen Zhang}, \bibinfo{person}{Yanwei Li}, \bibinfo{person}{Ziwei Liu}, {and} \bibinfo{person}{Chunyuan Li}.} \bibinfo{year}{2024}\natexlab{d}.
\newblock \showarticletitle{LLaVA-OneVision: Easy Visual Task Transfer}.
\newblock \bibinfo{journal}{\emph{CoRR}}  \bibinfo{volume}{abs/2408.03326} (\bibinfo{year}{2024}).
\newblock
\href{https://doi.org/10.48550/ARXIV.2408.03326}{doi:\nolinkurl{10.48550/ARXIV.2408.03326}}
\showeprint[arXiv]{2408.03326}


\bibitem[Li et~al\mbox{.}(2024a)]%
        {mike}
\bibfield{author}{\bibinfo{person}{Jiaqi Li}, \bibinfo{person}{Miaozeng Du}, \bibinfo{person}{Chuanyi Zhang}, \bibinfo{person}{Yongrui Chen}, \bibinfo{person}{Nan Hu}, \bibinfo{person}{Guilin Qi}, \bibinfo{person}{Haiyun Jiang}, \bibinfo{person}{Siyuan Cheng}, {and} \bibinfo{person}{Bozhong Tian}.} \bibinfo{year}{2024}\natexlab{a}.
\newblock \showarticletitle{{MIKE:} {A} New Benchmark for Fine-grained Multimodal Entity Knowledge Editing}. In \bibinfo{booktitle}{\emph{Findings of the Association for Computational Linguistics, {ACL} 2024, Bangkok, Thailand and virtual meeting, August 11-16, 2024}}, \bibfield{editor}{\bibinfo{person}{Lun{-}Wei Ku}, \bibinfo{person}{Andre Martins}, {and} \bibinfo{person}{Vivek Srikumar}} (Eds.). \bibinfo{publisher}{Association for Computational Linguistics}, \bibinfo{pages}{5018--5029}.
\newblock
\href{https://doi.org/10.18653/V1/2024.FINDINGS-ACL.298}{doi:\nolinkurl{10.18653/V1/2024.FINDINGS-ACL.298}}


\bibitem[Li et~al\mbox{.}(2024b)]%
        {codalm}
\bibfield{author}{\bibinfo{person}{Yanze Li}, \bibinfo{person}{Wenhua Zhang}, \bibinfo{person}{Kai Chen}, \bibinfo{person}{Yanxin Liu}, \bibinfo{person}{Pengxiang Li}, \bibinfo{person}{Ruiyuan Gao}, \bibinfo{person}{Lanqing Hong}, \bibinfo{person}{Meng Tian}, \bibinfo{person}{Xinhai Zhao}, \bibinfo{person}{Zhenguo Li}, {et~al\mbox{.}}} \bibinfo{year}{2024}\natexlab{b}.
\newblock \showarticletitle{Automated Evaluation of Large Vision-Language Models on Self-driving Corner Cases}.
\newblock \bibinfo{journal}{\emph{arXiv preprint arXiv:2404.10595}} (\bibinfo{year}{2024}).
\newblock


\bibitem[Li et~al\mbox{.}(2024c)]%
        {CODA-LM}
\bibfield{author}{\bibinfo{person}{Yanze Li}, \bibinfo{person}{Wenhua Zhang}, \bibinfo{person}{Kai Chen}, \bibinfo{person}{Yanxin Liu}, \bibinfo{person}{Pengxiang Li}, \bibinfo{person}{Ruiyuan Gao}, \bibinfo{person}{Lanqing Hong}, \bibinfo{person}{Meng Tian}, \bibinfo{person}{Xinhai Zhao}, \bibinfo{person}{Zhenguo Li}, \bibinfo{person}{Dit{-}Yan Yeung}, \bibinfo{person}{Huchuan Lu}, {and} \bibinfo{person}{Xu Jia}.} \bibinfo{year}{2024}\natexlab{c}.
\newblock \showarticletitle{Automated Evaluation of Large Vision-Language Models on Self-driving Corner Cases}.
\newblock \bibinfo{journal}{\emph{CoRR}}  \bibinfo{volume}{abs/2404.10595} (\bibinfo{year}{2024}).
\newblock
\href{https://doi.org/10.48550/ARXIV.2404.10595}{doi:\nolinkurl{10.48550/ARXIV.2404.10595}}
\showeprint[arXiv]{2404.10595}


\bibitem[Liu et~al\mbox{.}(2024a)]%
        {liu2024codeupdatearena}
\bibfield{author}{\bibinfo{person}{Zeyu~Leo Liu}, \bibinfo{person}{Shrey Pandit}, \bibinfo{person}{Xi Ye}, \bibinfo{person}{Eunsol Choi}, {and} \bibinfo{person}{Greg Durrett}.} \bibinfo{year}{2024}\natexlab{a}.
\newblock \showarticletitle{Codeupdatearena: Benchmarking knowledge editing on api updates}.
\newblock \bibinfo{journal}{\emph{arXiv preprint arXiv:2407.06249}} (\bibinfo{year}{2024}).
\newblock


\bibitem[Liu et~al\mbox{.}(2024b)]%
        {DBLP:journals/corr/abs-2407-06249}
\bibfield{author}{\bibinfo{person}{Zeyu~Leo Liu}, \bibinfo{person}{Shrey Pandit}, \bibinfo{person}{Xi Ye}, \bibinfo{person}{Eunsol Choi}, {and} \bibinfo{person}{Greg Durrett}.} \bibinfo{year}{2024}\natexlab{b}.
\newblock \showarticletitle{CodeUpdateArena: Benchmarking Knowledge Editing on {API} Updates}.
\newblock \bibinfo{journal}{\emph{CoRR}}  \bibinfo{volume}{abs/2407.06249} (\bibinfo{year}{2024}).
\newblock
\href{https://doi.org/10.48550/ARXIV.2407.06249}{doi:\nolinkurl{10.48550/ARXIV.2407.06249}}
\showeprint[arXiv]{2407.06249}


\bibitem[Ma et~al\mbox{.}(2024)]%
        {comprehendedit}
\bibfield{author}{\bibinfo{person}{Yaohui Ma}, \bibinfo{person}{Xiaopeng Hong}, \bibinfo{person}{Shizhou Zhang}, \bibinfo{person}{Huiyun Li}, \bibinfo{person}{Zhilin Zhu}, \bibinfo{person}{Wei Luo}, {and} \bibinfo{person}{Zhiheng Ma}.} \bibinfo{year}{2024}\natexlab{}.
\newblock \showarticletitle{ComprehendEdit: {A} Comprehensive Dataset and Evaluation Framework for Multimodal Knowledge Editing}.
\newblock \bibinfo{journal}{\emph{CoRR}}  \bibinfo{volume}{abs/2412.12821} (\bibinfo{year}{2024}).
\newblock
\href{https://doi.org/10.48550/ARXIV.2412.12821}{doi:\nolinkurl{10.48550/ARXIV.2412.12821}}
\showeprint[arXiv]{2412.12821}


\bibitem[Marcu et~al\mbox{.}(2024)]%
        {lingoqa}
\bibfield{author}{\bibinfo{person}{Ana{-}Maria Marcu}, \bibinfo{person}{Long Chen}, \bibinfo{person}{Jan H{\"{u}}nermann}, \bibinfo{person}{Alice Karnsund}, \bibinfo{person}{Beno{\^{\i}}t Hanotte}, \bibinfo{person}{Prajwal Chidananda}, \bibinfo{person}{Saurabh Nair}, \bibinfo{person}{Vijay Badrinarayanan}, \bibinfo{person}{Alex Kendall}, \bibinfo{person}{Jamie Shotton}, \bibinfo{person}{Elahe Arani}, {and} \bibinfo{person}{Oleg Sinavski}.} \bibinfo{year}{2024}\natexlab{}.
\newblock \showarticletitle{LingoQA: Visual Question Answering for Autonomous Driving}. In \bibinfo{booktitle}{\emph{Computer Vision - {ECCV} 2024 - 18th European Conference, Milan, Italy, September 29-October 4, 2024, Proceedings, Part {LXXVII}}} \emph{(\bibinfo{series}{Lecture Notes in Computer Science}, Vol.~\bibinfo{volume}{15135})}, \bibfield{editor}{\bibinfo{person}{Ales Leonardis}, \bibinfo{person}{Elisa Ricci}, \bibinfo{person}{Stefan Roth}, \bibinfo{person}{Olga Russakovsky}, \bibinfo{person}{Torsten Sattler}, {and} \bibinfo{person}{G{\"{u}}l Varol}} (Eds.). \bibinfo{publisher}{Springer}, \bibinfo{pages}{252--269}.
\newblock
\href{https://doi.org/10.1007/978-3-031-72980-5\_15}{doi:\nolinkurl{10.1007/978-3-031-72980-5\_15}}


\bibitem[Meng et~al\mbox{.}(2022)]%
        {rome}
\bibfield{author}{\bibinfo{person}{Kevin Meng}, \bibinfo{person}{David Bau}, \bibinfo{person}{Alex Andonian}, {and} \bibinfo{person}{Yonatan Belinkov}.} \bibinfo{year}{2022}\natexlab{}.
\newblock \showarticletitle{Locating and Editing Factual Associations in {GPT}}. In \bibinfo{booktitle}{\emph{Advances in Neural Information Processing Systems 35: Annual Conference on Neural Information Processing Systems 2022, NeurIPS 2022, New Orleans, LA, USA, November 28 - December 9, 2022}}, \bibfield{editor}{\bibinfo{person}{Sanmi Koyejo}, \bibinfo{person}{S.~Mohamed}, \bibinfo{person}{A.~Agarwal}, \bibinfo{person}{Danielle Belgrave}, \bibinfo{person}{K.~Cho}, {and} \bibinfo{person}{A.~Oh}} (Eds.).
\newblock
\urldef\tempurl%
\url{http://papers.nips.cc/paper\_files/paper/2022/hash/6f1d43d5a82a37e89b0665b33bf3a182-Abstract-Conference.html}
\showURL{%
\tempurl}


\bibitem[Meng et~al\mbox{.}(2023)]%
        {memit}
\bibfield{author}{\bibinfo{person}{Kevin Meng}, \bibinfo{person}{Arnab~Sen Sharma}, \bibinfo{person}{Alex~J. Andonian}, \bibinfo{person}{Yonatan Belinkov}, {and} \bibinfo{person}{David Bau}.} \bibinfo{year}{2023}\natexlab{}.
\newblock \showarticletitle{Mass-Editing Memory in a Transformer}. In \bibinfo{booktitle}{\emph{The Eleventh International Conference on Learning Representations, {ICLR} 2023, Kigali, Rwanda, May 1-5, 2023}}. \bibinfo{publisher}{OpenReview.net}.
\newblock
\urldef\tempurl%
\url{https://openreview.net/forum?id=MkbcAHIYgyS}
\showURL{%
\tempurl}


\bibitem[Mitchell et~al\mbox{.}(2022)]%
        {mend}
\bibfield{author}{\bibinfo{person}{Eric Mitchell}, \bibinfo{person}{Charles Lin}, \bibinfo{person}{Antoine Bosselut}, \bibinfo{person}{Chelsea Finn}, {and} \bibinfo{person}{Christopher~D. Manning}.} \bibinfo{year}{2022}\natexlab{}.
\newblock \showarticletitle{Fast Model Editing at Scale}. In \bibinfo{booktitle}{\emph{The Tenth International Conference on Learning Representations, {ICLR} 2022, Virtual Event, April 25-29, 2022}}. \bibinfo{publisher}{OpenReview.net}.
\newblock
\urldef\tempurl%
\url{https://openreview.net/forum?id=0DcZxeWfOPt}
\showURL{%
\tempurl}


\bibitem[Qian et~al\mbox{.}(2024)]%
        {NuScenes-QA}
\bibfield{author}{\bibinfo{person}{Tianwen Qian}, \bibinfo{person}{Jingjing Chen}, \bibinfo{person}{Linhai Zhuo}, \bibinfo{person}{Yang Jiao}, {and} \bibinfo{person}{Yu{-}Gang Jiang}.} \bibinfo{year}{2024}\natexlab{}.
\newblock \showarticletitle{NuScenes-QA: {A} Multi-Modal Visual Question Answering Benchmark for Autonomous Driving Scenario}. In \bibinfo{booktitle}{\emph{Thirty-Eighth {AAAI} Conference on Artificial Intelligence, {AAAI} 2024, Thirty-Sixth Conference on Innovative Applications of Artificial Intelligence, {IAAI} 2024, Fourteenth Symposium on Educational Advances in Artificial Intelligence, {EAAI} 2014, February 20-27, 2024, Vancouver, Canada}}, \bibfield{editor}{\bibinfo{person}{Michael~J. Wooldridge}, \bibinfo{person}{Jennifer~G. Dy}, {and} \bibinfo{person}{Sriraam Natarajan}} (Eds.). \bibinfo{publisher}{{AAAI} Press}, \bibinfo{pages}{4542--4550}.
\newblock
\href{https://doi.org/10.1609/AAAI.V38I5.28253}{doi:\nolinkurl{10.1609/AAAI.V38I5.28253}}


\bibitem[Sima et~al\mbox{.}(2024)]%
        {drivelm}
\bibfield{author}{\bibinfo{person}{Chonghao Sima}, \bibinfo{person}{Katrin Renz}, \bibinfo{person}{Kashyap Chitta}, \bibinfo{person}{Li Chen}, \bibinfo{person}{Hanxue Zhang}, \bibinfo{person}{Chengen Xie}, \bibinfo{person}{Jens Bei{\ss}wenger}, \bibinfo{person}{Ping Luo}, \bibinfo{person}{Andreas Geiger}, {and} \bibinfo{person}{Hongyang Li}.} \bibinfo{year}{2024}\natexlab{}.
\newblock \showarticletitle{DriveLM: Driving with Graph Visual Question Answering}. In \bibinfo{booktitle}{\emph{Computer Vision - {ECCV} 2024 - 18th European Conference, Milan, Italy, September 29-October 4, 2024, Proceedings, Part {LII}}} \emph{(\bibinfo{series}{Lecture Notes in Computer Science}, Vol.~\bibinfo{volume}{15110})}, \bibfield{editor}{\bibinfo{person}{Ales Leonardis}, \bibinfo{person}{Elisa Ricci}, \bibinfo{person}{Stefan Roth}, \bibinfo{person}{Olga Russakovsky}, \bibinfo{person}{Torsten Sattler}, {and} \bibinfo{person}{G{\"{u}}l Varol}} (Eds.). \bibinfo{publisher}{Springer}, \bibinfo{pages}{256--274}.
\newblock
\href{https://doi.org/10.1007/978-3-031-72943-0\_15}{doi:\nolinkurl{10.1007/978-3-031-72943-0\_15}}


\bibitem[Team(2025)]%
        {qwen2.5-VL}
\bibfield{author}{\bibinfo{person}{Qwen Team}.} \bibinfo{year}{2025}\natexlab{}.
\newblock \bibinfo{title}{Qwen2.5-VL}.
\newblock
\urldef\tempurl%
\url{https://qwenlm.github.io/blog/qwen2.5-vl/}
\showURL{%
\tempurl}


\bibitem[Wang et~al\mbox{.}(2024a)]%
        {qwen2vl}
\bibfield{author}{\bibinfo{person}{Peng Wang}, \bibinfo{person}{Shuai Bai}, \bibinfo{person}{Sinan Tan}, \bibinfo{person}{Shijie Wang}, \bibinfo{person}{Zhihao Fan}, \bibinfo{person}{Jinze Bai}, \bibinfo{person}{Keqin Chen}, \bibinfo{person}{Xuejing Liu}, \bibinfo{person}{Jialin Wang}, \bibinfo{person}{Wenbin Ge}, \bibinfo{person}{Yang Fan}, \bibinfo{person}{Kai Dang}, \bibinfo{person}{Mengfei Du}, \bibinfo{person}{Xuancheng Ren}, \bibinfo{person}{Rui Men}, \bibinfo{person}{Dayiheng Liu}, \bibinfo{person}{Chang Zhou}, \bibinfo{person}{Jingren Zhou}, {and} \bibinfo{person}{Junyang Lin}.} \bibinfo{year}{2024}\natexlab{a}.
\newblock \showarticletitle{Qwen2-VL: Enhancing Vision-Language Model's Perception of the World at Any Resolution}.
\newblock \bibinfo{journal}{\emph{CoRR}}  \bibinfo{volume}{abs/2409.12191} (\bibinfo{year}{2024}).
\newblock
\href{https://doi.org/10.48550/ARXIV.2409.12191}{doi:\nolinkurl{10.48550/ARXIV.2409.12191}}
\showeprint[arXiv]{2409.12191}


\bibitem[Wang et~al\mbox{.}(2024c)]%
        {wise}
\bibfield{author}{\bibinfo{person}{Peng Wang}, \bibinfo{person}{Zexi Li}, \bibinfo{person}{Ningyu Zhang}, \bibinfo{person}{Ziwen Xu}, \bibinfo{person}{Yunzhi Yao}, \bibinfo{person}{Yong Jiang}, \bibinfo{person}{Pengjun Xie}, \bibinfo{person}{Fei Huang}, {and} \bibinfo{person}{Huajun Chen}.} \bibinfo{year}{2024}\natexlab{c}.
\newblock \showarticletitle{{WISE}: Rethinking the Knowledge Memory for Lifelong Model Editing of Large Language Models}. In \bibinfo{booktitle}{\emph{The Thirty-eighth Annual Conference on Neural Information Processing Systems}}.
\newblock
\urldef\tempurl%
\url{https://openreview.net/forum?id=VJMYOfJVC2}
\showURL{%
\tempurl}


\bibitem[Wang et~al\mbox{.}(2024b)]%
        {DBLP:journals/corr/abs-2401-10471}
\bibfield{author}{\bibinfo{person}{Yiwei Wang}, \bibinfo{person}{Muhao Chen}, \bibinfo{person}{Nanyun Peng}, {and} \bibinfo{person}{Kai{-}Wei Chang}.} \bibinfo{year}{2024}\natexlab{b}.
\newblock \showarticletitle{DeepEdit: Knowledge Editing as Decoding with Constraints}.
\newblock \bibinfo{journal}{\emph{CoRR}}  \bibinfo{volume}{abs/2401.10471} (\bibinfo{year}{2024}).
\newblock
\href{https://doi.org/10.48550/ARXIV.2401.10471}{doi:\nolinkurl{10.48550/ARXIV.2401.10471}}
\showeprint[arXiv]{2401.10471}


\bibitem[Wu et~al\mbox{.}(2024)]%
        {DBLP:conf/emnlp/WuPWL24}
\bibfield{author}{\bibinfo{person}{Xiaobao Wu}, \bibinfo{person}{Liangming Pan}, \bibinfo{person}{William~Yang Wang}, {and} \bibinfo{person}{Anh~Tuan Luu}.} \bibinfo{year}{2024}\natexlab{}.
\newblock \showarticletitle{{AKEW:} Assessing Knowledge Editing in the Wild}. In \bibinfo{booktitle}{\emph{Proceedings of the 2024 Conference on Empirical Methods in Natural Language Processing, {EMNLP} 2024, Miami, FL, USA, November 12-16, 2024}}, \bibfield{editor}{\bibinfo{person}{Yaser Al{-}Onaizan}, \bibinfo{person}{Mohit Bansal}, {and} \bibinfo{person}{Yun{-}Nung Chen}} (Eds.). \bibinfo{publisher}{Association for Computational Linguistics}, \bibinfo{pages}{15118--15133}.
\newblock
\urldef\tempurl%
\url{https://aclanthology.org/2024.emnlp-main.843}
\showURL{%
\tempurl}


\bibitem[Xing et~al\mbox{.}(2024)]%
        {openemma}
\bibfield{author}{\bibinfo{person}{Shuo Xing}, \bibinfo{person}{Chengyuan Qian}, \bibinfo{person}{Yuping Wang}, \bibinfo{person}{Hongyuan Hua}, \bibinfo{person}{Kexin Tian}, \bibinfo{person}{Yang Zhou}, {and} \bibinfo{person}{Zhengzhong Tu}.} \bibinfo{year}{2024}\natexlab{}.
\newblock \showarticletitle{OpenEMMA: Open-Source Multimodal Model for End-to-End Autonomous Driving}.
\newblock \bibinfo{journal}{\emph{CoRR}}  \bibinfo{volume}{abs/2412.15208} (\bibinfo{year}{2024}).
\newblock
\href{https://doi.org/10.48550/ARXIV.2412.15208}{doi:\nolinkurl{10.48550/ARXIV.2412.15208}}
\showeprint[arXiv]{2412.15208}


\bibitem[Xu et~al\mbox{.}(2024)]%
        {DBLP:conf/cikm/XuZZLL00WY0C024}
\bibfield{author}{\bibinfo{person}{Derong Xu}, \bibinfo{person}{Ziheng Zhang}, \bibinfo{person}{Zhihong Zhu}, \bibinfo{person}{Zhenxi Lin}, \bibinfo{person}{Qidong Liu}, \bibinfo{person}{Xian Wu}, \bibinfo{person}{Tong Xu}, \bibinfo{person}{Wanyu Wang}, \bibinfo{person}{Yuyang Ye}, \bibinfo{person}{Xiangyu Zhao}, \bibinfo{person}{Enhong Chen}, {and} \bibinfo{person}{Yefeng Zheng}.} \bibinfo{year}{2024}\natexlab{}.
\newblock \showarticletitle{Editing Factual Knowledge and Explanatory Ability of Medical Large Language Models}. In \bibinfo{booktitle}{\emph{Proceedings of the 33rd {ACM} International Conference on Information and Knowledge Management, {CIKM} 2024, Boise, ID, USA, October 21-25, 2024}}, \bibfield{editor}{\bibinfo{person}{Edoardo Serra} {and} \bibinfo{person}{Francesca Spezzano}} (Eds.). \bibinfo{publisher}{{ACM}}, \bibinfo{pages}{2660--2670}.
\newblock
\href{https://doi.org/10.1145/3627673.3679673}{doi:\nolinkurl{10.1145/3627673.3679673}}


\bibitem[Yang et~al\mbox{.}(2025)]%
        {realworldedit}
\bibfield{author}{\bibinfo{person}{Wanli Yang}, \bibinfo{person}{Fei Sun}, \bibinfo{person}{Jiajun Tan}, \bibinfo{person}{Xinyu Ma}, \bibinfo{person}{Qi Cao}, \bibinfo{person}{Dawei Yin}, \bibinfo{person}{Huawei Shen}, {and} \bibinfo{person}{Xueqi Cheng}.} \bibinfo{year}{2025}\natexlab{}.
\newblock \showarticletitle{The Mirage of Model Editing: Revisiting Evaluation in the Wild}.
\newblock \bibinfo{journal}{\emph{CoRR}}  \bibinfo{volume}{abs/2502.11177} (\bibinfo{year}{2025}).
\newblock
\href{https://doi.org/10.48550/ARXIV.2502.11177}{doi:\nolinkurl{10.48550/ARXIV.2502.11177}}
\showeprint[arXiv]{2502.11177}


\bibitem[Yao et~al\mbox{.}(2024)]%
        {calmm}
\bibfield{author}{\bibinfo{person}{Ruoyu Yao}, \bibinfo{person}{Yubin Wang}, \bibinfo{person}{Haichao Liu}, \bibinfo{person}{Rui Yang}, \bibinfo{person}{Zengqi Peng}, \bibinfo{person}{Lei Zhu}, {and} \bibinfo{person}{Jun Ma}.} \bibinfo{year}{2024}\natexlab{}.
\newblock \showarticletitle{CALMM-Drive: Confidence-Aware Autonomous Driving with Large Multimodal Model}.
\newblock \bibinfo{journal}{\emph{CoRR}}  \bibinfo{volume}{abs/2412.04209} (\bibinfo{year}{2024}).
\newblock
\href{https://doi.org/10.48550/ARXIV.2412.04209}{doi:\nolinkurl{10.48550/ARXIV.2412.04209}}
\showeprint[arXiv]{2412.04209}


\bibitem[Yao et~al\mbox{.}(2025)]%
        {yao2025cake}
\bibfield{author}{\bibinfo{person}{Yunzhi Yao}, \bibinfo{person}{Jizhan Fang}, \bibinfo{person}{Jia-Chen Gu}, \bibinfo{person}{Ningyu Zhang}, \bibinfo{person}{Shumin Deng}, \bibinfo{person}{Huajun Chen}, {and} \bibinfo{person}{Nanyun Peng}.} \bibinfo{year}{2025}\natexlab{}.
\newblock \showarticletitle{CaKE: Circuit-aware Editing Enables Generalizable Knowledge Learners}.
\newblock \bibinfo{journal}{\emph{arXiv preprint arXiv:2503.16356}} (\bibinfo{year}{2025}).
\newblock


\bibitem[Yao et~al\mbox{.}(2023)]%
        {yao2023editing}
\bibfield{author}{\bibinfo{person}{Yunzhi Yao}, \bibinfo{person}{Peng Wang}, \bibinfo{person}{Bozhong Tian}, \bibinfo{person}{Siyuan Cheng}, \bibinfo{person}{Zhoubo Li}, \bibinfo{person}{Shumin Deng}, \bibinfo{person}{Huajun Chen}, {and} \bibinfo{person}{Ningyu Zhang}.} \bibinfo{year}{2023}\natexlab{}.
\newblock \showarticletitle{Editing Large Language Models: Problems, Methods, and Opportunities}. In \bibinfo{booktitle}{\emph{Proceedings of the 2023 Conference on Empirical Methods in Natural Language Processing, {EMNLP} 2023, Singapore, December 6-10, 2023}}, \bibfield{editor}{\bibinfo{person}{Houda Bouamor}, \bibinfo{person}{Juan Pino}, {and} \bibinfo{person}{Kalika Bali}} (Eds.). \bibinfo{publisher}{Association for Computational Linguistics}, \bibinfo{pages}{10222--10240}.
\newblock
\href{https://doi.org/10.18653/V1/2023.EMNLP-MAIN.632}{doi:\nolinkurl{10.18653/V1/2023.EMNLP-MAIN.632}}


\bibitem[Youssef et~al\mbox{.}(2025)]%
        {youssef2025position}
\bibfield{author}{\bibinfo{person}{Paul Youssef}, \bibinfo{person}{Zhixue Zhao}, \bibinfo{person}{Daniel Braun}, \bibinfo{person}{J{\"o}rg Schl{\"o}tterer}, {and} \bibinfo{person}{Christin Seifert}.} \bibinfo{year}{2025}\natexlab{}.
\newblock \showarticletitle{Position: Editing Large Language Models Poses Serious Safety Risks}.
\newblock \bibinfo{journal}{\emph{arXiv preprint arXiv:2502.02958}} (\bibinfo{year}{2025}).
\newblock


\bibitem[Zhang et~al\mbox{.}(2024c)]%
        {mcmke}
\bibfield{author}{\bibinfo{person}{Junzhe Zhang}, \bibinfo{person}{Huixuan Zhang}, \bibinfo{person}{Xunjian Yin}, \bibinfo{person}{Baizhou Huang}, \bibinfo{person}{Xu Zhang}, \bibinfo{person}{Xinyu Hu}, {and} \bibinfo{person}{Xiaojun Wan}.} \bibinfo{year}{2024}\natexlab{c}.
\newblock \showarticletitle{{MC-MKE:} {A} Fine-Grained Multimodal Knowledge Editing Benchmark Emphasizing Modality Consistency}.
\newblock \bibinfo{journal}{\emph{CoRR}}  \bibinfo{volume}{abs/2406.13219} (\bibinfo{year}{2024}).
\newblock
\href{https://doi.org/10.48550/ARXIV.2406.13219}{doi:\nolinkurl{10.48550/ARXIV.2406.13219}}
\showeprint[arXiv]{2406.13219}


\bibitem[Zhang et~al\mbox{.}(2024a)]%
        {instructedit}
\bibfield{author}{\bibinfo{person}{Ningyu Zhang}, \bibinfo{person}{Bozhong Tian}, \bibinfo{person}{Siyuan Cheng}, \bibinfo{person}{Xiaozhuan Liang}, \bibinfo{person}{Yi Hu}, \bibinfo{person}{Kouying Xue}, \bibinfo{person}{Yanjie Gou}, \bibinfo{person}{Xi Chen}, {and} \bibinfo{person}{Huajun Chen}.} \bibinfo{year}{2024}\natexlab{a}.
\newblock \showarticletitle{InstructEdit: Instruction-Based Knowledge Editing for Large Language Models}. In \bibinfo{booktitle}{\emph{Proceedings of the Thirty-Third International Joint Conference on Artificial Intelligence, {IJCAI} 2024, Jeju, South Korea, August 3-9, 2024}}. \bibinfo{publisher}{ijcai.org}, \bibinfo{pages}{6633--6641}.
\newblock
\urldef\tempurl%
\url{https://www.ijcai.org/proceedings/2024/733}
\showURL{%
\tempurl}


\bibitem[Zhang et~al\mbox{.}(2024b)]%
        {comprehensive}
\bibfield{author}{\bibinfo{person}{Ningyu Zhang}, \bibinfo{person}{Yunzhi Yao}, \bibinfo{person}{Bozhong Tian}, \bibinfo{person}{Peng Wang}, \bibinfo{person}{Shumin Deng}, \bibinfo{person}{Mengru Wang}, \bibinfo{person}{Zekun Xi}, \bibinfo{person}{Shengyu Mao}, \bibinfo{person}{Jintian Zhang}, \bibinfo{person}{Yuansheng Ni}, \bibinfo{person}{Siyuan Cheng}, \bibinfo{person}{Ziwen Xu}, \bibinfo{person}{Xin Xu}, \bibinfo{person}{Jia{-}Chen Gu}, \bibinfo{person}{Yong Jiang}, \bibinfo{person}{Pengjun Xie}, \bibinfo{person}{Fei Huang}, \bibinfo{person}{Lei Liang}, \bibinfo{person}{Zhiqiang Zhang}, \bibinfo{person}{Xiaowei Zhu}, \bibinfo{person}{Jun Zhou}, {and} \bibinfo{person}{Huajun Chen}.} \bibinfo{year}{2024}\natexlab{b}.
\newblock \showarticletitle{A Comprehensive Study of Knowledge Editing for Large Language Models}.
\newblock \bibinfo{journal}{\emph{CoRR}}  \bibinfo{volume}{abs/2401.01286} (\bibinfo{year}{2024}).
\newblock
\href{https://doi.org/10.48550/ARXIV.2401.01286}{doi:\nolinkurl{10.48550/ARXIV.2401.01286}}
\showeprint[arXiv]{2401.01286}


\bibitem[Zhang et~al\mbox{.}(2023)]%
        {adalora}
\bibfield{author}{\bibinfo{person}{Qingru Zhang}, \bibinfo{person}{Minshuo Chen}, \bibinfo{person}{Alexander Bukharin}, \bibinfo{person}{Pengcheng He}, \bibinfo{person}{Yu Cheng}, \bibinfo{person}{Weizhu Chen}, {and} \bibinfo{person}{Tuo Zhao}.} \bibinfo{year}{2023}\natexlab{}.
\newblock \showarticletitle{Adaptive Budget Allocation for Parameter-Efficient Fine-Tuning}. In \bibinfo{booktitle}{\emph{The Eleventh International Conference on Learning Representations}}.
\newblock
\urldef\tempurl%
\url{https://openreview.net/forum?id=lq62uWRJjiY}
\showURL{%
\tempurl}


\bibitem[Zheng et~al\mbox{.}(2023)]%
        {ike}
\bibfield{author}{\bibinfo{person}{Ce Zheng}, \bibinfo{person}{Lei Li}, \bibinfo{person}{Qingxiu Dong}, \bibinfo{person}{Yuxuan Fan}, \bibinfo{person}{Zhiyong Wu}, \bibinfo{person}{Jingjing Xu}, {and} \bibinfo{person}{Baobao Chang}.} \bibinfo{year}{2023}\natexlab{}.
\newblock \showarticletitle{Can We Edit Factual Knowledge by In-Context Learning?}. In \bibinfo{booktitle}{\emph{Proceedings of the 2023 Conference on Empirical Methods in Natural Language Processing, {EMNLP} 2023, Singapore, December 6-10, 2023}}, \bibfield{editor}{\bibinfo{person}{Houda Bouamor}, \bibinfo{person}{Juan Pino}, {and} \bibinfo{person}{Kalika Bali}} (Eds.). \bibinfo{publisher}{Association for Computational Linguistics}, \bibinfo{pages}{4862--4876}.
\newblock
\href{https://doi.org/10.18653/V1/2023.EMNLP-MAIN.296}{doi:\nolinkurl{10.18653/V1/2023.EMNLP-MAIN.296}}


\end{thebibliography}

\end{document}